\documentclass{article}
\PassOptionsToPackage{usenames,svgnames,dvipsnames}{xcolor}

 \usepackage[main,nonatbib,final]{neurips_2026}
\usepackage{amsmath,amsfonts,bm}

\newcommand{\newterm}[1]{{\bf #1}}

\def\Figref#1{Figure~\ref{#1}}

\def\eqref#1{equation~\ref{#1}}
\def\1{\bm{1}}
\newcommand{\train}{\mathcal{D}}

\def\vone{{\bm{1}}}

\def\vtheta{{\bm{\theta}}}

\def\vx{{\bm{x}}}

\def\mA{{\bm{A}}}
\def\mB{{\bm{B}}}

\def\mH{{\bm{H}}}
\def\mI{{\bm{I}}}

\def\mK{{\bm{K}}}
\def\mL{{\bm{L}}}

\DeclareMathAlphabet{\mathsfit}{\encodingdefault}{\sfdefault}{m}{sl}
\SetMathAlphabet{\mathsfit}{bold}{\encodingdefault}{\sfdefault}{bx}{n}

\def\sS{{\mathbb{S}}}

\def\sX{{\mathbb{X}}}

\newcommand{\E}{\mathbb{E}}
\newcommand{\Ls}{\mathcal{L}}

\usepackage[utf8]{inputenc} % allow utf-8 input
\usepackage[T1]{fontenc}    % use 8-bit T1 fonts
\usepackage{hyperref}
\usepackage{url}
\usepackage{graphicx}
\usepackage{booktabs}
\usepackage[normalem]{ulem}
\usepackage{soul}
\usepackage{adjustbox}
\usepackage{amssymb}
\usepackage{algorithm}
\usepackage{algpseudocode}
\usepackage{paralist}
\usepackage{multirow}
\usepackage[font=small]{caption}
\usepackage{csquotes}
\usepackage{wrapfig}
\usepackage{multirow}
\usepackage{textcomp}
\usepackage{mathtools}
\usepackage{bm}
\usepackage{tabularx,array}
\usepackage[numbers]{natbib}
\usepackage[capitalize,noabbrev]{cleveref}
\usepackage{xcolor}

\definecolor{lightblue}{rgb}{0.88, 0.95, 0.98}
\definecolor{lightpink}{rgb}{0.98, 0.88, 0.95}
\definecolor{superlightred}{rgb}{0.99, 0.92, 0.92}
\definecolor{superlightgreen}{rgb}{0.92, 0.99, 0.92}
\definecolor{darkergreen}{rgb}{0.1, 0.6, 0.1}
\definecolor{squareblue}{HTML}{6ce3cf}
\usepackage{algorithm}
\usepackage{algpseudocode}
\usepackage[table]{xcolor}

\hypersetup{%
  pdfborder = {0 0 0},
  bookmarksdepth = section,
  citecolor = DarkGreen,
  linkcolor = DarkBlue,
  urlcolor = DarkBlue,
  colorlinks = true
}%

\title{Network of Theseus (Like the ship)}

\author{Vighnesh Subramaniam\textsuperscript{1}\thanks{Corresponding author. Code: \url{https://github.com/vsubramaniam851/net-convert}}~, Colin Conwell\textsuperscript{2}, \textbf{Boris Katz}\textsuperscript{1}\textbf{,}
\\\textbf{Andrei Barbu}\textsuperscript{1}\textbf{,} \textbf{Brian Cheung}\textsuperscript{3}\\
$^1$MIT CSAIL $^2$Department of Cognitive Science, Johns Hopkins University $^3$UCSF
\\
$^{1}$\texttt{\small\{vsub851,boris,abarbu\}@mit.edu} \\$^2$\texttt{\small cconwel2@jhu.edu}\\ $^3$\texttt{\small bcheung@ucsf.edu}
}

\begin{document}

\maketitle

\begin{abstract}
A standard assumption in deep learning is that the inductive bias introduced by a neural network architecture must persist from training through inference. The architecture you train with is the architecture you deploy. This assumption constrains the community from selecting architectures that may have desirable efficiency or design properties due to difficulties with optimization. We challenge this assumption with the Network of Theseus (NoT), a method for progressively converting a trained, or even untrained, guide network architecture part-by-part into an entirely different target network architecture while preserving the performance of the guide network. At each stage, components in the guide network architecture are incrementally replaced with target architecture modules and aligned via representational similarity metrics. This procedure preserves most of the functionality of the guide network even under substantial architectural changes—for example, converting a convolutional network into a multilayer perceptron, or GPT-2 into a recurrent neural network. By decoupling optimization from deployment, NoT expands the space of viable inference-time architectures, opening opportunities for better accuracy–efficiency tradeoffs and enabling more directed exploration of the architectural design space.

\end{abstract}

\section{Introduction}
In machine learning research, we tend to assume that training is coupled with inference. If a specific architecture is trained to solve a task with specific inductive biases and computational mechanisms, that same architecture should be used at test time. This assumption is embedded in common practice. Neural architecture search (NAS) discovers an efficient inference-time architecture that is then trained and deployed \citep{zoph2016neural, real2019regularized, liu2018darts}. Compression pipelines prune and quantize a trained model to meet deployment budgets while preserving the trained structure \citep{han2015deep, jacob2018quantization}. This assumption applies even in settings like distillation, where teacher networks train student networks designed explicitly for inference \citep{hinton2015distilling, gou2021knowledge, romero2014fitnets, huang2022knowledge}. A few lines of work partially relax this coupling by training a supernet and selecting subnets for deployment \citep{cai2019once, yu2018slimmable}, but the resulting inference models remain architectural subgraphs or variants of the original training network rather than cross-architectural conversions.

We challenge this premise that the architecture used for training must be the architecture used for inference. Decoupling the architecture used for training from that used for inference would enable models to be trained with large, optimization friendly architectures and converted into lighter architectures for efficient inference on edge devices. This would also enable controlled exploration of inductive biases by comparing architectures without confounding optimization difficulty.

Large-scale analyses demonstrate that very different architectures (e.g. CNNs vs Transformers) converge toward similar internal representations as they scale \citep{huh2024platonic, han2023system, li2015convergent, conwell2024large}. Furthermore, classic observations of representational alignment across independently trained networks \citep{raghu2017svcca, kornblith2019similarity} suggest that sufficiently expressive, distinct architectures can implement the same input–output functions.  This statement of universality is also supported by theoretical work that shows that functions that are Turing-computable can be approximated by any neural network \citep{poggiofraser2024}. However, the real challenge is not whether such functions exist, but how to reach them through optimization. This view implies that architectural priors are not constraints that must persist at inference, but are primarily training biases -- scaffolds designed to guide and stabilize optimization.

Motivated by these perspectives and potential benefits, we introduce \emph{Network of Theseus} (NoT), a part-by-part conversion procedure that starts from a given guide architecture and progressively replaces its components (e.g. layers) with target modules. Our name for this procedure is a reference to Plutarch’s Ship of Theseus paradox \citep{plutarch_parallel_lives}, which asks: once all the decaying planks of a ship are replaced (and none of the original remain), is it still the same ship?

At each replacement stage, we instantiate a target module and optimize it to match the activations of the guide network using a representational distance function \citep{kornblith2019similarity, cristianini2001kernel, cortes2012algorithms}. This optimization transfers priors between networks at a lower level, e.g. at the layer level rather than at the architectural level \citep{subramaniam2025training}. After the conversion is complete, the resultant target architecture is trained end-to-end on the downstream task. This staged alignment decouples training from deployment: we can train with one architecture and convert it into an entirely different one for test-time use, aiming to preserve the original performance as closely as possible.

Using NoT, we show broad architectural conversions including convolutional (CNN) models to fully connected MLPs with low-rank linear layers, vision transformers with multihead attention to token-wise MLPs, and transformer language models to Elman RNNs. Interestingly, we observe even untrained guide networks contain useful inductive biases; NoT transfers these effectively, yielding performance comparable to using trained guides. Furthermore, we find that NoT can convert from larger to smaller networks and that converting from a larger to a smaller architecture (e.g., a deeper ResNet to a shallower one) can at minimum preserve performance. We show our conversions using centered kernel alignment (CKA) and further validate our findings by introducing a new kernel-based similarity metric, differentiable mutual nearest neighbors (D-MNN). Designed from the metric introduced in \citet{huh2024platonic}, D-MNN exploits local geometric structure for comparing representations.

NoT has both practical and theoretical implications for architectural transfer and design. With trained or untrained guides, we can convert to architectures that are more efficient or better aligned with deployment constraints or theoretical desiderata. Because NoT aims to preserve performance during conversion, it enables discovery of inference architectures with accuracy–efficiency trade-offs not previously attained, including transfers from larger (even untrained) guides to smaller targets. NoT implies that designing and training an architecture with simple optimization dynamics is not necessary: we can now design and deploy neural networks where constraints matter, expanding architectural choice at inference time without resolving the full supervised learning problem from
scratch for every candidate design. 

Our contributions are as follows:

\begin{compactenum}
    \item We introduce the Network of Theseus (NoT), a method to part-by-part convert from one architecture to another using representational alignment.
    \item We demonstrate NoT across a wide variety of architectural conversions such as converting convolutions in ResNet-18 to linear layers to create an MLP or converting attention in GPT-2 to RNN layers to create a Deep RNN. We find that across all conversions, we preserve performance of the original architecture despite dramatic architectural shifts.
    \item Surprisingly, we find that we can perform the same conversion starting from an untrained network architecture, demonstrating that we can transfer inductive biases to our target networks. 
    \item We validate our findings using representational similarity metrics like CKA and introduce a new metric based on a differentiable version of mutual nearest neighbors, D-MNN, to further validate our findings as general and invariant to metric. 
\end{compactenum}

\begin{figure*}[t]
    \centering
    \includegraphics[width=\textwidth]{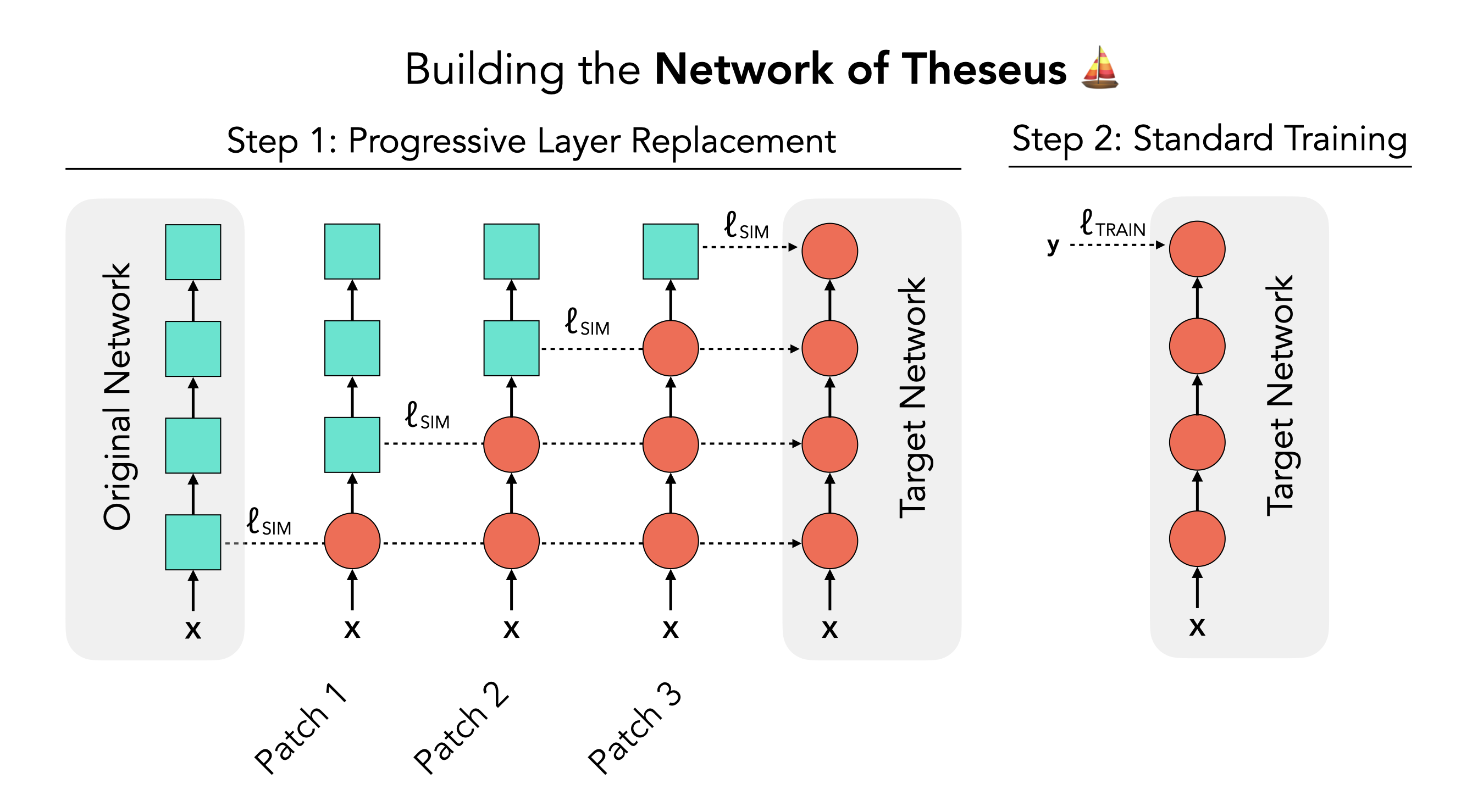}
    \caption{\textbf{Network of Theseus}. A network can be converted to any desired target network by replacing each piece of the original network incrementally, part-by-part. Each original part is replaced by optimizing the representational alignment, $\ell_{SIM}$, of the target part to the original part. After all original parts are replaced, only the target network remains and can be trained on any downstream task (i.e. standard training).}
    \label{fig:prog_method}
    \vspace{-2ex}
\end{figure*}
\vspace{-2ex}

\section{Related Work}
\vspace{-1ex}
\textbf{Architectural Transfer via Linearizing Transformers}: NoT is a general method that builds on work applying cross-architectural distillation for linearizing transformers \citep{mercat2024linearizing, zhang2024lolcats, zhang2024hedgehog, bick2024transformers}. These works focus on specifically distilling multihead attention to either SSMs \citep{wang2024mamba, bick2025llamba} or linear attention. Most approaches involve training linear attention layers or SSMs to approximate softmax attention via MSE. Some show that linearizing can specifically scale to very large transformers \citep{zhang2024lolcats} e.g. 7B - 72B LLMs without considerable effect on performance. 

These papers are significantly different from NoT, both in scope and approach. We greatly expand architectural transfer with linear attention or SSMs to transfers that are not dependent on having the same computations in order to use MSE. Our representational distance functions allow for any architectural pairing as well, making our conversions much richer. While previous work linearized with only linear attention, we can now focus on other layer replacements such as converting convolutional layers to linear layers, etc. NoT also makes methodological changes to how layer swaps are applied, considering changes that are staged and progressive rather than all-at-once. Our comparisons show that this progressive replacement is important, especially when architectures are different.

\textbf{Model Distillation and Compression}: Distillation \citep{hinton2015distilling, gou2021knowledge, sanh2019distilbert, hsieh2023distilling, tian2019contrastive, chen2021wasserstein, lin2020ensemble} transfers knowledge from a teacher model to a student model by introducing a new component to the loss function that forces the student model to behave like the teacher model \citep{kim2021comparing, zhou2021bert}. This usually consists of penalizing the KL-divergence between the logit predictions of the student and teacher model. Methods have been proposed that use CKA as an alignment approach between representations of two networks or with representations in the brain with notable improvement in network performance \citep{saha2022distilling, dapello2022aligning}. 

Other works also transfer knowledge across models while compressing information. For example, BERT-of-Theseus \citep{xu-etal-2020-bert} uses the "Ship-of-Theseus" framing to compress BERT into a smaller architecture using stochastic replacement schedule during supervised training. NoT significantly expands on BERT-of-Theseus to include multiple replacement schedules, a wider range of architectures, and untrained starting architectures. Similarly, other papers gradually convert layers to identity functions in order to shrink the given architecture \citep{chen2023vanillanetpowerminimalismdeep}, where adjacent convolutions are merged for inference. This uses a within‑architecture training trick to prune non‑linearities and fuse layers. 

We distinguish NoT from distillation and compression. NoT aligns computations and can transfer inductive biases from one architecture to another. Like prior work \citep{subramaniam2025training, ulyanov2018deep, zhong2024algorithmic}, this allows us to use completely untrained guide networks as part of our conversion, although our replacement strategy in NoT is distinct from this previous work. Doing the same with distillation would lead to much worse results \citep{subramaniam2025training}. NoT is significantly different from previous work using feature kernels to convert across architectures \citep{subramaniam2025training, he2022feature}. To our knowledge, the stage-wise replacement of NoT has not been proposed in previous work. These works do not use representational alignment as an initialization method as well.

\textbf{Representational Convergence}: Our paper builds on prior investigations into representational alignment across different network architectures. Prior work has seen both behavioral and representational convergence across diverse neural network architectures. For example, prior work has found that networks trained with self-supervised and supervised have similar errors, indicating behavioral alignment \citep{geirhos2020surprising}. Many other prior works have discovered similar representational alignment through different similarity metrics such as kernel alignment \citep{huh2024platonic, han2023system} and model stitching \citep{bansal2021revisiting}. These prior works have discovered representational convergence across different architectures, learning tasks, and even modalities. This representational convergence across architectures and learning tasks have been exploited in prior work to improve network performance. For example, representational alignment is used to fix architectures that are traditionally ill-suited for certain tasks due to optimization failures or diffusion models with inconsistent representations \citep{yu2024representation}. Similar to these, we use representational convergence in NoT to convert across architectures progressively.

\section{Methods: Network of Theseus (NoT)}
\label{sec:method}
\newterm{Network of Theseus (NoT)} provides a general procedure for replacing parts of a source (``guide'') network with parts of a target architecture while preserving internal representations. At any time, a subset of components is replaced; we then train the replaced subset to \emph{match} the guide’s activations, stage by stage, according to a \newterm{replacement schedule}. \Figref{fig:prog_method} illustrates our default schedule, \newterm{progressive replacement}, which we find works best empirically. \Figref{fig:replacement_methods} compares alternative schedules. NoT is task-agnostic. After the replacement completes, we finetune end-to-end for the downstream objective (e.g., image classification or language modeling).

Let $f^G(\cdot)$ be a fixed guide network with $k$ layers and let $f^T(\cdot;\theta)$ denote the under-construction target network obtained by replacing some of the guide’s layers with new modules (e.g., Conv2d$\rightarrow$Linear, attention variants, or block-level substitutions). We assume that both networks share input/output interfaces so that they can be run on the same inputs $\vx \in \sX$ . For a mini-batch from train, let $\mA_i^G(\vx)~\text{and}~\mA_i^T(\vx;\theta)$ be the activations extracted at layer $i$ from $f^G$ and $f^T$, respectively. In practice, for similarity computation, we flatten spatial/token dimensions. 

We write $g(\cdot, \cdot) \in [0, 1]$ for a representational similarity (e.g., linear CKA) and use its complement  $\Delta(\mA, \mB) \triangleq 1 - g(\mA, \mB) \in [0, 1]$ as a \newterm{dissimilarity} to be minimized.

\subsection{Part-by-part matching objective}
\label{sec:replace}
For illustrative purposes, we show how to apply NoT for a single layer $i$. The matching loss is $\Ls_i(\vtheta_i)~=~\E_{\vx\sim\train}\left[\Delta\!\left(\mA_i^T(\vx;\vtheta_i),\mA_i^G(\vx)\right)\right]$
where $\theta_i$ are the parameters introduced at location $i$ by the replacement. At a training \emph{stage}, we optimize a set $I\subseteq\{1, \cdots,k\}$ of replaced layers jointly:

\begin{equation}
\label{eq:stage}
\Ls_{I}(\vtheta_{I})~=~\E_{\vx\sim\train}\left[\frac{1}{|I|}\sum_{i\in I}\Delta\!\left(\mA_i^T(\vx;\vtheta_{I}), \mA_i^G(\vx)\right)\right].
\end{equation}
Layers not in $I$ are frozen to prevent drift in the guide pathway.

\textbf{Replacement Modules}: NoT is agnostic to the specific replacement, requiring only shape compatibility. For Conv2d, we use an \emph{equivalent low-rank linear layer} that applies to a flattened input and reshapes back to the expected output. Because of a large parameter count for the target linear layer, we first apply a linear layer to project the input to a lower dimensionality and then use another linear layer to project the desired output dimensionality, both of which are tunable. For attention, we support replacements with basic tokenwise MLPs, without any token-mixing, or RNNs. The framework also accommodates block-level replacements e.g., ResNet-50$\rightarrow$ ResNet-18 basic blocks.

\subsection{Replacement Schedules}
\label{sec:schedule}
\begin{figure}[t]
    \centering
    \includegraphics[width=0.8\textwidth]{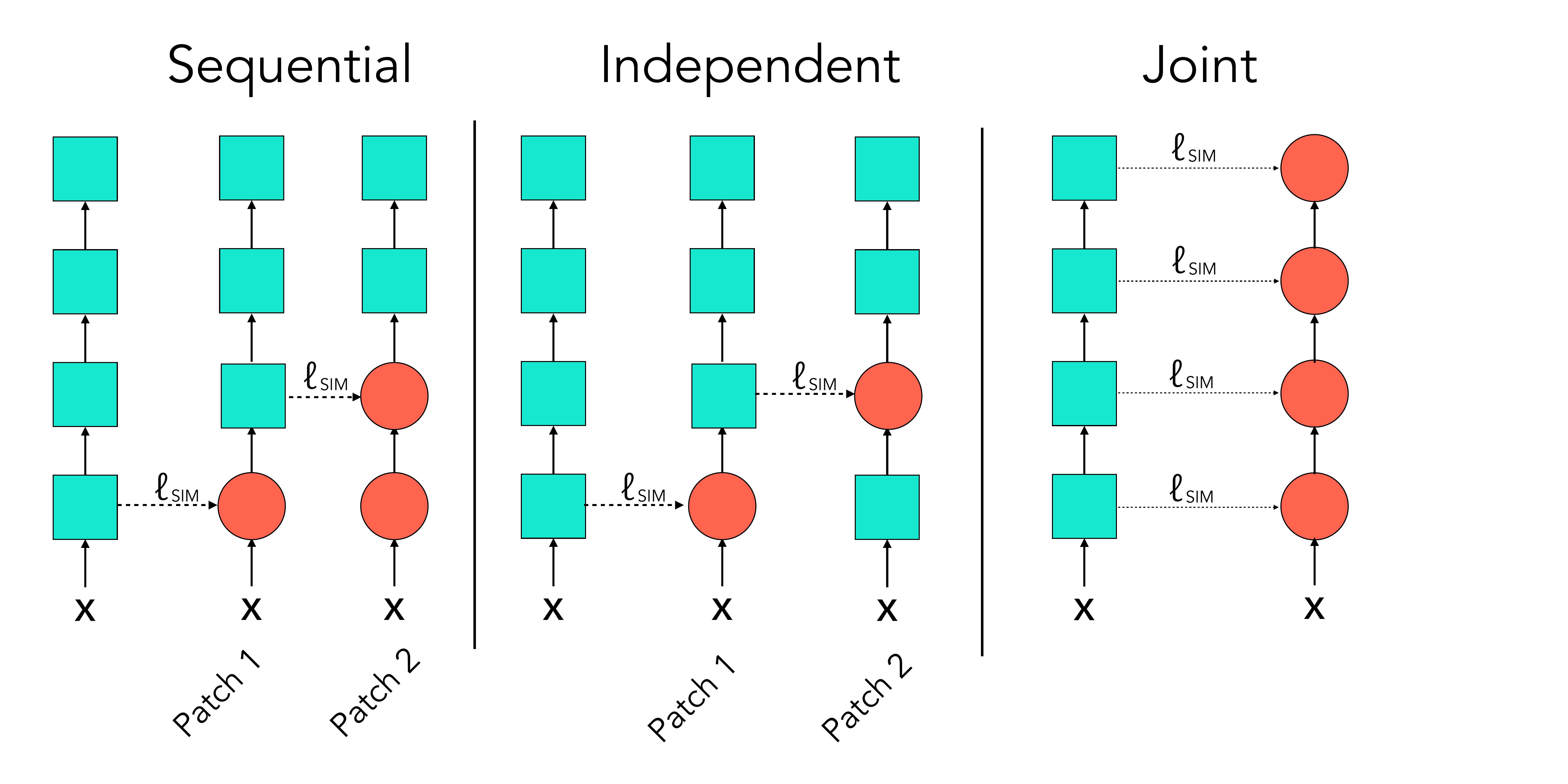}
    \caption{\textbf{Alternative replacement schedules}: Sequential freezes earlier replacements. Independent fits each layer using guide inputs. Joint trains all target layers at once.}
    \label{fig:replacement_methods}
    \vspace{-3ex}
\end{figure}

A \newterm{replacement schedule} is a sequence $\sS = (I_1, \cdots, I_T)$ of index sets indicating which layers are trainable at each stage. NoT supports several schedules; we adopt \textbf{progressive replacement} by default due to its favorable stability–accuracy tradeoff.

\textbf{Progressive}: At stage $t$, we replace the $t$-th layer and set $I_t = \{1, \cdots t\}$ jointly optimizing all previously replaced layers together with the new one. This minimizes error accumulation while x feature distributions calibrated across the replaced prefix (\Cref{fig:prog_method}). 

\textbf{Sequential}: Identical to progressive in the order of replacement, but we \emph{freeze} all previously replaced layers and optimize only the newest one, i.e., $I_t = \{t\}$ with earlier replacements fixed. This can suffer from error accumulation because misalignment in earlier layers propagates forward (\Cref{fig:replacement_methods}).

\textbf{Independent}: Each layer $i$ is optimized \emph{in place} using the guide’s input distribution at that layer, without coupling to other replacements. After fitting all layers independently, we assemble the full target. This avoids sequential error accumulation but introduces distribution shift once the independently trained layers are composed (\Cref{fig:replacement_methods}).

\textbf{Joint}: Replace all $k$ layers at once and set $I_1=\{1,\cdots,k\}$. This removes schedule bias but can be harder to optimize because later layers must adapt to rapidly changing early layers (\Cref{fig:replacement_methods}). This is closest to \citep{subramaniam2025training} although our application is closest to an initialization.

\textbf{Group-wise variants}: Beyond scalar layer indices, $I_t$ can denote \emph{groups} (e.g., a ResNet group). Our approach supports progressive group introduction (e.g., ResNet-50 $\rightarrow$ ResNet-18) as well as progressive \emph{alignment} where a smaller student unfreezes groups over time to match a fixed teacher.

\vspace{-2ex}
\subsection{Similarity choices and Tasks}
\label{sec:similarity}
We instantiate $g$ with \newterm{linear CKA} or \newterm{differentiable mutual nearest neighbors (D-MNN)}, both of which are scale-invariant across rows (mini-batch samples). D-MNN is a representational similarity metric designed and introduced in this paper based on the nearest neighbor metric used in \citet{huh2024platonic}. In this work, we were inspired by the locality property of the metric and aimed to make it differentiable in the hope that local structure in the representation space has structure that can be exploited and aligned. We refer to \Cref{ap:cka} and \Cref{ap:dknn}. 
\vspace{-2ex}

\section{Experiments}
\label{sec:exp}
\begin{figure}
    \centering
    \includegraphics[width=\textwidth]{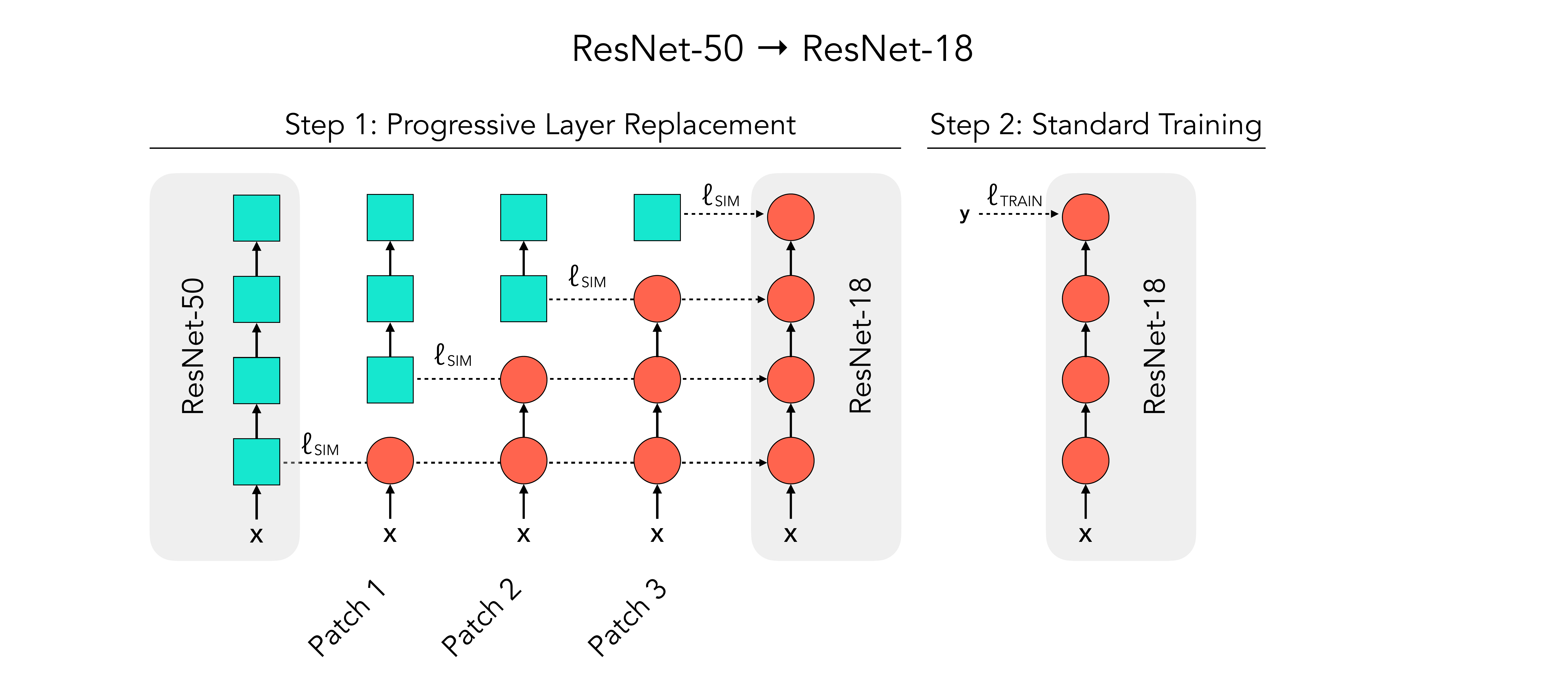}
    \caption{\textbf{NoT makes any-to-any architecture conversion possible.} Without the requirement of functional intermediate models, the target network can be any architecture. For instance, we convert the deeper ResNet-50 to the shallower ResNet-18. Blue squares are a set of multiple ResNet-50 blocks. These are replaced with a single ResNet-18 block. ResNet-18 blocks are added until we have a full ResNet-18 network.}
    \label{fig:resnet18_theseus}
\end{figure}

We apply NoT across a range of architectures and replacements to measure how well we can use our staged replacement across different replacement schedules. For all experiments, we compare using a trained and untrained guide network and separate training for representational similarity and task training into two stages. 

\textbf{Tasks}: We consider image classification and language modeling in this paper. For testing image classification, we use the ImageNet dataset \citep{deng2009imagenet}, measuring Top-1 performance on the pre-defined validation set. For language modeling, we use the Wikitext-103 dataset \citep{merity2016pointer} where models must predict the next token given some context. We use a sequence length of 128 for all models and use the training, validation and testing sets defined by the dataset. We tokenize the text using the GPT-2 \citep{radford2019language} tokenizer.

\textbf{Architectures and Replacements}: We study cross-family and within-family conversions spanning convolutions, attention, and recurrent computation. In all cases, layers are progressively replaced and optimized to match the guide’s intermediate activations via representational similarity.

\textit{ResNet-18$\rightarrow$MLP}: We convert ResNet-18 \citep{he2016deep} to a fully-connected MLP by replacing each convolutional layer with a linear layer. This removes the spatial priors of convolutional layers such as receptive fields or translational equivariance. We use two low-rank linear layers (rank 256-1024) to keep the parameter size of the linear layers reasonable, as discussed in \Cref{sec:replace}. Batch normalization recalibration details are in \Cref{ap:methods}.

\textit{DINOv2$\rightarrow$Patch-MLP}: We convert DINOv2 \citep{oquab2023dinov2} to a patch-wise MLP by replacing each multihead attention with a token-wise MLP that applies feedforward layers independently to each visual token. This eliminates cross-token communication, increasing efficiency at the cost of token interaction.

\textit{GPT-2$\rightarrow$RNN}: We replace GPT-2 \citep{radford2019language, vaswani2017attention} multihead attention layers with two Elman RNN layers per attention block. This tests whether sequential memory-based processing can substitute for parallel attention mechanisms.

\textit{ResNet-50$\rightarrow$ResNet-18}: We study architectural transfer rather than direct replacements that require matching input and output shapes. We map every four ResNet-50 blocks to one ResNet-18 block (blocks follow \citep{he2016deep}), see \Cref{fig:resnet18_theseus}. We rebuild ResNet-18 block by block, independent of the guide architecture. At each stage the new block learns to match the representations of its four source blocks, testing whether NoT transfers depth based expressivity to a shallower model. This setting shows NoT can operate across any architectural conversion. See \Cref{ap:rn50} for details.

\textit{GPT-2 Large$\rightarrow$GPT-2 Small}: We conduct the same architectural transfer for language models. We map every three GPT-2 Large blocks to one GPT-2 Small block (blocks follow \cite{radford2019language, vaswani2017attention} and correspond to transformer decoder layers). We rebuild GPT-2 block by block, independent of the guide architecture. At each stage, the new block learns to match the representations of its three blocks.

% \textit{ResNet-50$\rightarrow$ResNet-18}: We expand beyond direct replacement where our input-output shapes need to match to general architectural transfer. To do this, we convert from ResNet-50 to ResNet-18 by mapping every 4 ResNet-50 blocks to 1 ResNet-18 block (where blocks are defined in \citep{he2016deep}). See \Cref{fig:resnet18_theseus}. We make the progressive training architecture-agnostic by reconstructing ResNet-18 independently, block by block, without incorporating into the guide architecture. At each stage, a ResNet-18 block is optimized to match the representations of the corresponding 4 ResNet-50 blocks, testing whether NoT can transfer depth-based expressivity to shallower architectures. We cover more details in \Cref{ap:rn50}.

\textbf{Training}:
For each setting, we train NoT with a guide network and the target replacements. We include a baseline with \emph{naive replacement} where all guide network layers are replaced with the target modules from scratch for a fair comparison to understand what representational similarity provides to performance. The baseline is FLOP matched so this is trained with tasked training for the same number of optimization steps as progressive replacement, including alignment and standard training. All replacement schedules are also trained for the same total optimization steps as well, so they are FLOP-matched. During task training, all networks are trained with cross-entropy loss, without loss of generality. For all task training and representational similarity optimization, we use the AdamW \citep{loshchilov2017decoupled} optimizer. When optimizing representational similarity, we also incorporate gradient clipping due to unstable training. We use CKA and D-MNN for all settings. 

We use batch size 256 throughout because representational similarity estimates improve with more samples. Alignment learning rates and epochs are tuned per stage (15--100 epochs), and image-classification runs use warmup plus cosine scheduling \citep{loshchilov2016sgdr}. Task training runs for 100 epochs with swept learning rates and 5--10 warmup epochs under the same scheduler; full hyperparameters are in \Cref{ap:exp}. After tuning, each setting is run with 5 seeds and we report standard errors. For Wikitext-103, test perplexity is taken from the epoch with the lowest seed-averaged validation loss.
\vspace{-2ex}

\section{Results}

\begin{figure}
    \centering
    \includegraphics[width=\textwidth]{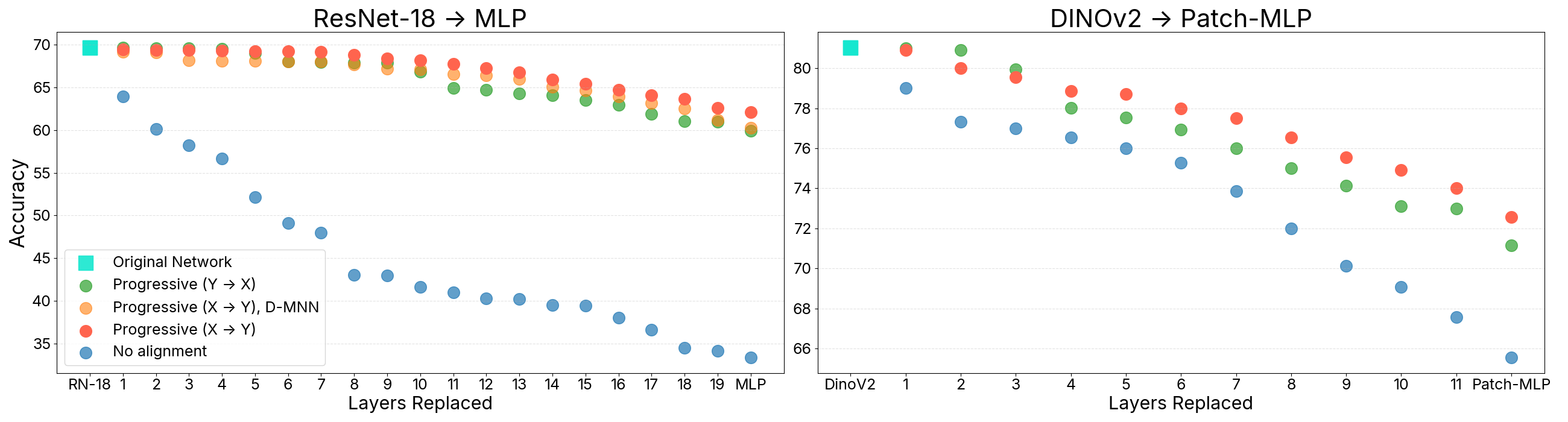}
    \caption{\small\textbf{Progressive layer replacement preserves performance.} Across patches, NoT aligns each replacement before finetuning the hybrid network, far exceeding naive replacement. Forward and reverse replacement perform similarly, and D-MNN is consistent with CKA in ResNet-18$\rightarrow$MLP.}

    \label{fig:layer_replace}
    \vspace{-1ex}
\end{figure}

\begin{table}[t]
\centering  
\begin{tabular}{lcccc}
\toprule
Original (Guide) $\rightarrow$ Target & Similarity Metric & Guide & NoT & Baseline \\
\midrule
& & \multicolumn{3}{c}{\textbf{ImageNet Top-1 Accuracy} ($\uparrow$)} \\
\cmidrule(lr){3-5}
ResNet-18 $\rightarrow$ MLP &  CKA   & 69.66 & 62.12$\pm$0.42 (\,$\uparrow$\,22.11) & 40.01$\pm$1.44\\
ResNet-18 $\rightarrow$ MLP & D-MNN  & 69.66 & 60.28$\pm$0.39 (\,$\uparrow$\,20.27) & 40.01$\pm$1.44 \\
DINOv2 $\rightarrow$ Patch-MLP & CKA & 81.03 & 72.56$\pm$0.20 (\,$\uparrow$\,3.84) & 68.72$\pm$0.55 \\
DINOv2 $\rightarrow$ Patch-MLP & D-MNN & 81.03 & 73.02$\pm$0.84 (\,$\uparrow$\,4.30) & 68.72$\pm$0.55 \\
ResNet-50 $\rightarrow$ ResNet-18 & CKA & 76.13 & 73.81$\pm$0.26 (\,$\uparrow$\,4.91) & 68.90$\pm$0.33 \\
ResNet-50 $\rightarrow$ ResNet-18 & D-MNN & 76.13 & 73.02 $\pm$0.84 (\,$\uparrow$4.12) & 68.90$\pm$0.33
\\
\addlinespace
& & \multicolumn{3}{c}{\textbf{Perplexity} ($\downarrow$)} \\
\cmidrule(lr){3-5}
GPT-2 $\rightarrow$ RNN & CKA & 37.50 & 50.58$\pm$0.89 (\,$\downarrow$\,68.86) & 119.44$\pm$3.01 \\
GPT-2 $\rightarrow$ RNN & D-MNN & 37.50 & 47.44$\pm$1.91 (\,$\downarrow$\,70.00) & 119.44$\pm$3.01 \\
GPT-2 Large $\rightarrow$ GPT-2 Small & UCKA & 22.05 & 24.98$\pm$1.52 (\,$\downarrow$\,3.03) & 28.01$\pm$1.02 \\ 
GPT-2 Large $\rightarrow$ GPT-2 Small & D-MNN & 22.05 & 26.02$\pm$1.22 (\,$\downarrow$\,1.99) & 28.01$\pm$1.02 \\ 
\bottomrule
\end{tabular}
\caption{\textbf{Network of Theseus vastly improves over naive replacement and preserves performance}. We compare NoT with a FLOP-matched standard training and the original guide network performance. We find that NoT vastly outperforms naive replacement, close to 25\% on ImageNet and 70 points on Wikitext-103.}
\label{tab:main}
\vspace{-4ex}
\end{table}

\begin{table}
\centering
\begin{adjustbox}{max width=\textwidth}
\begin{tabular}{lccccc}
\toprule
Original $\rightarrow$ Target & Similarity Metric & Progressive & Joint & Independent & Sequential \\
\midrule
& & \multicolumn{4}{c}{\textbf{ImageNet Top-1 Accuracy} ($\uparrow$)} \\
\cmidrule(lr){3-6}
ResNet-18 $\rightarrow$ MLP & CKA & 62.12$\pm$0.42 & 57.98$\pm$0.28 & 45.10$\pm$0.23 & 30.73$\pm$0.14 \\
ResNet-18 $\rightarrow$ MLP & D-MNN & 60.28$\pm$0.39 &  57.66$\pm$0.22 &  45.81$\pm$0.06 & 31.90$\pm$0.33 \\
DINOv2 $\rightarrow$ Patch-MLP & CKA & 72.56$\pm$0.20 & 70.42$\pm$0.33 & 66.02$\pm$
0.47 & 47.15$\pm$0.32 \\
DINOv2 $\rightarrow$ Patch-MLP & D-MNN & 73.02$\pm$0.84 & 71.69$\pm$0.94 & 69.88$\pm$0.13 & 51.94$\pm$1.55 \\
ResNet-50 $\rightarrow$ ResNet-18 & CKA & 73.81$\pm$0.26 & 70.38$\pm$0.35 & 61.51$\pm$0.46 & 61.08$\pm$0.15\\
ResNet-50 $\rightarrow$ ResNet-18 & D-MNN & 73.02 $\pm$0.84 & 69.12$\pm$1.10 & 60.87$\pm$0.11 & 62.04$\pm$0.92\\
\addlinespace
& & \multicolumn{4}{c}{\textbf{Perplexity} ($\downarrow$)} \\
\cmidrule(lr){3-6}
GPT-2 $\rightarrow$ RNN & CKA & 50.58$\pm$0.89  & 57.69$\pm$0.58 & 100.94$\pm$1.15 & 121.36$\pm$2.23\\
GPT-2 $\rightarrow$ RNN & D-MNN & 47.44$\pm$1.91  & 56.66$\pm$1.55 & 115.96$\pm$1.65 & 136.98$\pm$0.54\\
GPT-2 Large $\rightarrow$ GPT-2 & UCKA & 24.98$\pm$1.52  & 25.04$\pm$1.33 & 33.66$\pm$0.94 & 29.62$\pm$2.36\\
GPT-2 Large $\rightarrow$ GPT-2 & D-MNN & 26.02$\pm$1.22  & 27.20$\pm$1.43 & 33.96$\pm$2.65 & 34.91$\pm$3.41\\
\bottomrule
\end{tabular}
\end{adjustbox}
\caption{\textbf{Progressive replacement outperforms other replacement schedules}: We compare progressive replacement with previously discussed replacement schedules, finding that our staged, progressive schedule is extremely useful. All replacement schedules are trained for the same FLOP budget. Joint replacement is the closest but likely requires significantly longer training.}
\label{tab:replace_comp}
\end{table}

\begin{table}[t]
\centering
\small
\setlength{\tabcolsep}{3pt}        % tighter columns
\renewcommand{\arraystretch}{1.05} % slightly tighter rows
\resizebox{0.9\columnwidth}{!}{%
\begin{tabular}{lcccc}
\toprule
\multirow{2}{*}{Original $\rightarrow$ Target} & \multirow{2}{*}{Similarity Metric} &
\multicolumn{2}{c}{\textbf{Guide}} & \multirow{2}{*}{Baseline} \\
\cmidrule(lr){3-4}
 & & Trained & Untrained & \\
\midrule
 & & \multicolumn{3}{c}{\textbf{ImageNet Top-1 Accuracy} ($\uparrow$)} \\
\cmidrule(lr){3-5}
ResNet-18 $\rightarrow$ MLP & CKA   & 62.12$\pm$0.42 & 60.85$\pm$0.48 & 40.01$\pm$1.44 \\
ResNet-18 $\rightarrow$ MLP & D-MNN & 60.28$\pm$0.39 & 60.90$\pm$0.69 & 40.01$\pm$1.44 \\
DINOv2 $\rightarrow$ Patch-MLP & CKA & 72.56$\pm$0.20 & 70.39$\pm$0.25 & 68.72$\pm$0.55 \\
DINOv2 $\rightarrow$ Patch-MLP & D-MNN & 73.02$\pm$0.84 & 68.15$\pm$0.96 & 68.72$\pm$0.55 \\
ResNet-50 $\rightarrow$ ResNet-18 & CKA & 73.81$\pm$0.26 & 71.61$\pm$0.42 & 68.90$\pm$0.33 \\
ResNet-50 $\rightarrow$ ResNet-18 & D-MNN & 73.02$\pm$0.84 & 70.22$\pm$1.18 & 68.90$\pm$0.33 \\
\addlinespace
 & & \multicolumn{3}{c}{\textbf{Wikitext Perplexity} ($\downarrow$)} \\
\cmidrule(lr){3-5}
GPT-2 $\rightarrow$ RNN & CKA & 50.58$\pm$0.89 & 58.26$\pm$0.43 & 119.44$\pm$3.01 \\
GPT-2 $\rightarrow$ RNN & D-MNN & 47.44$\pm$1.91 & 49.71$\pm$2.56 & 119.44$\pm$3.01 \\
GPT-2 Large $\rightarrow$ GPT-2 & UCKA & 24.98$\pm$1.52 & 25.10$\pm$1.97 & 28.01$\pm$1.02 \\
GPT-2 Large $\rightarrow$ GPT-2 & D-MNN & 26.02$\pm$1.22 & 26.15$\pm$2.43 & 28.01$\pm$1.02 \\
\bottomrule
\end{tabular}%
}
\caption{\textbf{NoT with untrained guide networks improves over naive replacement}. Can we apply NoT with untrained guide networks to transfer inductive bias? We find that untrained guide networks contain useful inductive biases and can improve over naive replacement with no alignment, given by baseline. NoT with untrained guide networks is competitive with NoT with trained guide networks.}
\label{tab:untrained}
\vspace{-5ex}
\end{table}

\textbf{NoT significantly outperforms naive replacement}: We apply NoT to all of our previously described networks and summarize the results in \Cref{tab:main}. We find that across all settings, NoT improves performance over naive replacement by up to 22\%. For example, we find that fully-connected MLPs can only be trained to achieve 40\% accuracy. With NoT, we improve by 22\% and identify a fully-connected MLP that is competitive with ResNet-18. Similarly, we are able to replace attention in vision transformers with Patch-MLPs while preserving accuracy, meaning that token communication is not necessary for downstream image classification, achieving competitive results on DINOv2. This holds for Elman RNNs, which become effective replacements for attention computations. Most excitingly, we find that NoT can be applied across similar architectures like ResNet-18. These results far exceed naive replacement and standard training. We also find consistent results with D-MNN as with CKA.

\begin{figure}
    \centering
    \includegraphics[width=0.9\textwidth]{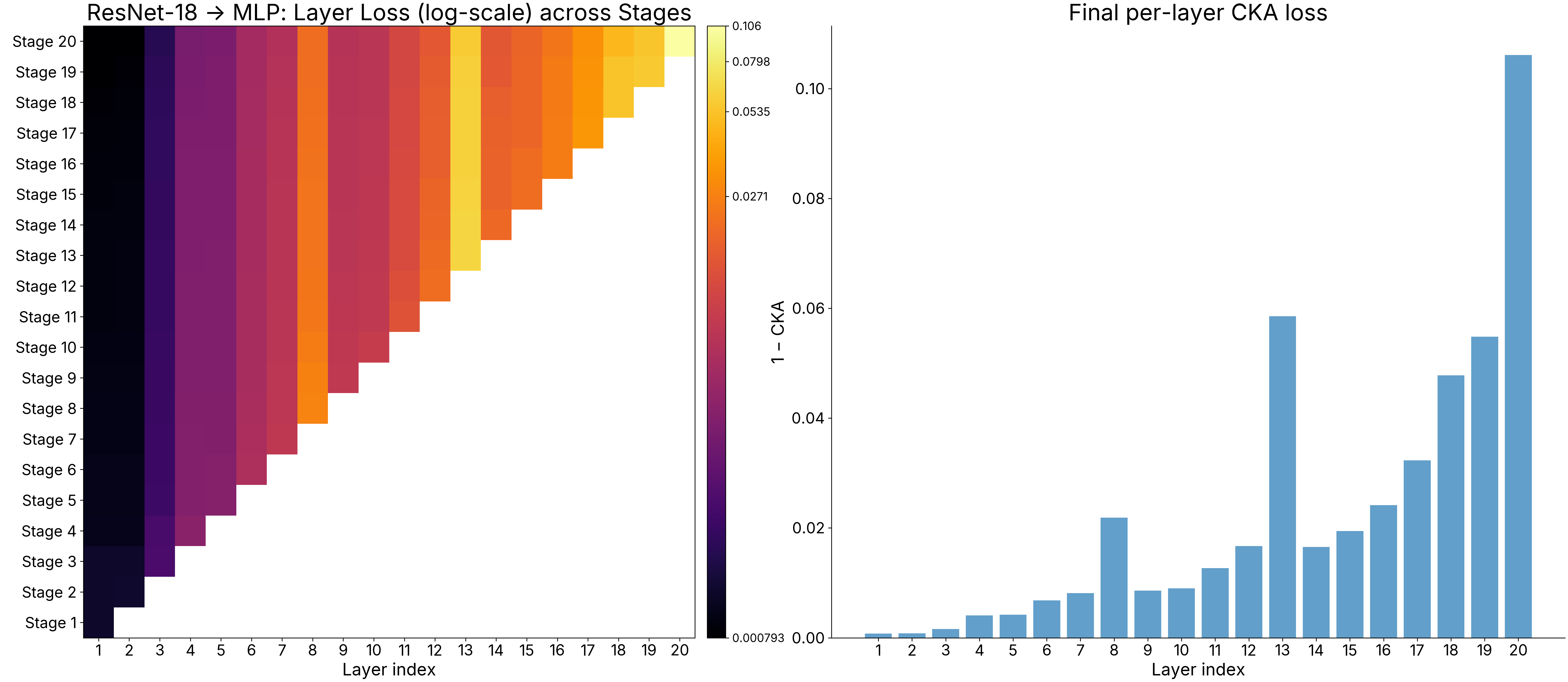}
    \caption{\textbf{Representational similarity across stages reveals difficult layers}: (left) We show CKA similarity losses (log-scaled) across all stages of progressive replacement. We see that CKA loss decreases across all stages for all layers. (right) We plot the final CKA loss for the last stage across all layers. We can identify bottleneck layers that are more difficult to align. Specifically, layers 6, 8, 13, and the final layers have higher loss. Layers 6, 8, and 13 are associated with downsampling in ResNet-18.}
    \label{fig:align_loss}
    \vspace{-4ex}
\end{figure}

\Cref{fig:layer_replace} tracks progressive replacement for ResNet-18 and DINOv2 against the guide and naive target. Forward and reverse schedules both outperform naive training. Reverse replacement is slightly weaker, likely from representational drift but still achieves strong performance. Drops at layers such as layer 13 of ResNet-18 indicate bottlenecks associated with downsampling layers.

\textbf{NoT significantly outperforms training from scratch}: Training MLPs on ImageNet remains difficult: \citet{bachmann2023scaling} report 51.5\% with heavy training and test-time augmentation. NoT achieves higher accuracy without such augmentation, highlighting the gap between functions an MLP can represent and those reached by direct optimization.

\textbf{Alignment becomes more challenging at the end}: \Cref{fig:align_loss} shows CKA loss during ResNet-18$\rightarrow$MLP alignment. Bottleneck layers such as 8 and 13, associated with downsampling in ResNet-18, are harder to align, and later layers are generally harder across reverse replacement, DINOv2, and GPT-2. Improved late-stage alignment is a clear target for future work.

\begin{figure}
    \vspace{-3ex}
    \centering
    \includegraphics[width=0.9\textwidth]{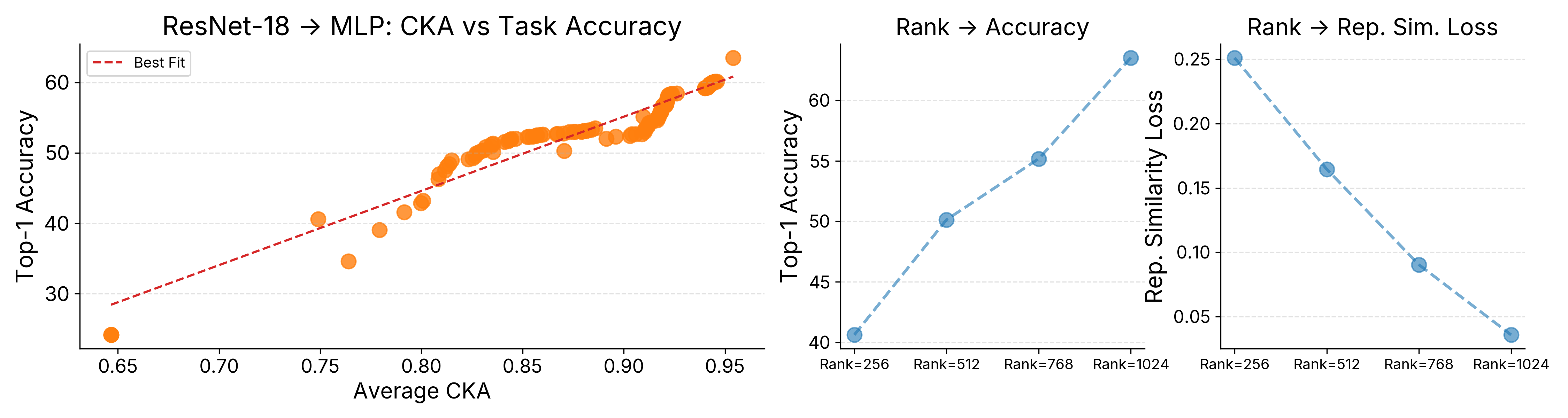}
    \caption{\textbf{NoT performance improves with larger CKA and larger ranks}: We compare final task performance of the ResNet-18$\rightarrow$MLP setting over different final CKA alignment scores (left) and linear layer ranks (right). We find that stronger average alignment for the final stage leads to better performance. Similarly, larger rank leads to better loss and stronger task performance.}
    \label{fig:analysis}
    \vspace{-2ex}
\end{figure}

\textbf{CKA alignment predicts accuracy and improves with rank}: Using checkpoints from the final progressive stage, \Cref{fig:analysis} shows that higher average CKA predicts higher task accuracy. Larger low-rank linear layers reduce CKA loss and improve downstream accuracy, suggesting further capacity or optimization can improve NoT.

\textbf{Progressive replacement outperforms other schedules}: \Cref{tab:replace_comp} compares progressive, joint, independent, and sequential schedules. Progressive is strongest across settings; joint is closest but likely needs more training, independent suffers composition-time distribution shift, and sequential accumulates error.

\textbf{Untrained guide networks improve over naive replacements}: \Cref{tab:untrained} shows that NoT transfers useful priors from untrained guides \citep{subramaniam2025training, ulyanov2018deep, zhong2024algorithmic}. This distinguishes NoT from distillation, which is weak with untrained teachers. For example, untrained ResNet-50$\rightarrow$ResNet-18 improves the target by 3\%, suggesting depth and connectivity can be transferable even without learned weights.

\section{Conclusion}
Training and inference are usually coupled, but Network of Theseus (NoT) breaks this coupling by progressively replacing a guide network with target modules while aligning intermediate representations; therefore, the trained function can be carried into a different inference architecture. Across ImageNet and Wikitext-103, this staged conversion preserves much of the guide’s performance for large cross-family changes, specifically ResNet-18$\rightarrow$MLP, DINOv2$\rightarrow$Patch-MLP, ResNet-50$\rightarrow$ResNet-18, GPT-2$\rightarrow$RNN, whereas naive replacement collapses. Therefore, representational alignment is the key mechanism.

Regardless of whether the guide architecture was trained for a downstream task or not, we find that such staged replacement allows for an inductive bias transfer, allowing us to discover target architectures that preserve performance in comparison to the guide architecture. Our results raise concrete scientific questions: What defines a “discoverable representation”, one that a different target architecture can reliably realize, and how can training objectives be designed to increase discoverability? Given that NoT decouples training from deployment, how should we jointly choose guide architectures (for favorable optimization geometry) and target architectures (for deployment desiderata such as latency, memory, and parallelism) to trace the optimal alignment–efficiency frontier?

We believe NoT reframes how we think about architectural design. If representational alignment can be used to carry performance across network architecture families, then the new goal of an architecture during training is to construct discoverable representations rather than dictating the deployed inference architecture.

\textbf{Limitations}: We have not fully explored the space of architectures in our current work and, more importantly, have not reached the boundary of the performance of our alignment procedure for two reasons. First, we have not reached a point of overfitting in \Cref{fig:analysis} that increased target network capacity continues to improve performance, indicating that further training is necessary to see such convergence. Second, this lack of coverage is due to computational limitations as the progressive process requires more resources that simply fitting all layers at once. This gap between our performance and the possible best performance is driven by certain bottleneck layers, e.g. optimizing the last layers as shown in \Cref{fig:align_loss}. Further training methods can be explored to improve these results such as more data augmentation.

In most cases, Network of Theseus does not drive target network to match or beat guide network performance. The remaining gap is likely driven by the capacity of the target network and what can learned. This gap is interesting because it is indicative of what is special about the current networks we are designing. Further research will be dedicated to tightening this gap.

\section*{Acknowledgments}
This work was supported by the DARPA Mathematics for the DIscovery of ALgorithms and Architectures (DIAL) program, the DARPA Knowledge Management at Scale and Speed (KMASS) program, the NSF award 2124052, the MIT-IBM Watson AI Lab, the Department of the Air Force Artificial Intelligence Accelerator under Cooperative Agreement Number FA8750-19-2-1000, and the Air Force Office of Scientific Research (AFOSR) under award number FA9550-21-1-0399. The views and conclusions contained in this document are those of the authors and should not be interpreted as representing the official policies, either expressed or implied, of the Department of the Air Force or the U.S. Government. The U.S. Government is authorized to reproduce and distribute reprints for Government purposes notwithstanding any copyright notation herein.

We would like to thank Chris Wang, David Mayo, Phillip Isola and the rest of the Isola lab for their broad feedback on this work.

\bibliographystyle{plainnat}
\bibliography{neurips_2026}

@article{tian2019contrastive,
  title={Contrastive representation distillation},
  author={Tian, Yonglong and Krishnan, Dilip and Isola, Phillip},
  journal={arXiv preprint arXiv:1910.10699},
  year={2019}
}

@inproceedings{kornblith2019similarity,
  title={Similarity of neural network representations revisited},
  author={Kornblith, Simon and Norouzi, Mohammad and Lee, Honglak and Hinton, Geoffrey},
  booktitle={International conference on machine learning},
  pages={3519--3529},
  year={2019},
  organization={PMLR}
}

@article{cristianini2001kernel,
  title={On kernel-target alignment},
  author={Cristianini, Nello and Shawe-Taylor, John and Elisseeff, Andre and Kandola, Jaz},
  journal={Advances in neural information processing systems},
  volume={14},
  year={2001}
}

@article{cortes2012algorithms,
  title={Algorithms for learning kernels based on centered alignment},
  author={Cortes, Corinna and Mohri, Mehryar and Rostamizadeh, Afshin},
  journal={The Journal of Machine Learning Research},
  volume={13},
  pages={795--828},
  year={2012},
  publisher={JMLR. org}
}

@inproceedings{he2016deep,
  title={Deep residual learning for image recognition},
  author={He, Kaiming and Zhang, Xiangyu and Ren, Shaoqing and Sun, Jian},
  booktitle={Proceedings of the IEEE conference on computer vision and pattern recognition},
  pages={770--778},
  year={2016}
}

@article{poggiofraser2024,
  title={Compositional Sparsity of Learnable Functions},
  author={Poggio, Tomaso and Fraser, Maia},
  year={2024},
journal={Bulletin of the American Mathematical Society}
}

@article{vaswani2017attention,
  title={Attention is all you need},
  author={Vaswani, A},
  journal={Advances in Neural Information Processing Systems},
  year={2017}
}

@inproceedings{han2023system,
  title={System identification of neural systems: If we got it right, would we know?},
  author={Han, Yena and Poggio, Tomaso A and Cheung, Brian},
  booktitle={International Conference on Machine Learning},
  pages={12430--12444},
  year={2023},
  organization={PMLR}
}

@inproceedings{deng2009imagenet,
  title={Imagenet: A large-scale hierarchical image database},
  author={Deng, Jia and Dong, Wei and Socher, Richard and Li, Li-Jia and Li, Kai and Fei-Fei, Li},
  booktitle={2009 IEEE conference on computer vision and pattern recognition},
  pages={248--255},
  year={2009},
  organization={Ieee}
}

@article{merity2016pointer,
  title={Pointer sentinel mixture models},
  author={Merity, Stephen and Xiong, Caiming and Bradbury, James and Socher, Richard},
  journal={arXiv preprint arXiv:1609.07843},
  year={2016}
}

@article{hinton2015distilling,
  title={Distilling the Knowledge in a Neural Network},
  author={Hinton, Geoffrey},
  journal={arXiv preprint arXiv:1503.02531},
  year={2015}
}

@article{gou2021knowledge,
  title={Knowledge distillation: A survey},
  author={Gou, Jianping and Yu, Baosheng and Maybank, Stephen J and Tao, Dacheng},
  journal={International Journal of Computer Vision},
  volume={129},
  number={6},
  pages={1789--1819},
  year={2021},
  publisher={Springer}
}

@article{sanh2019distilbert,
  title={DistilBERT, A Distilled Version of BERT: Smaller, Faster, Cheaper and Lighter},
  author={Sanh, V},
  journal={arXiv preprint arXiv:1910.01108},
  year={2019}
}

@article{hsieh2023distilling,
  title={Distilling step-by-step! outperforming larger language models with less training data and smaller model sizes},
  author={Hsieh, Cheng-Yu and Li, Chun-Liang and Yeh, Chih-Kuan and Nakhost, Hootan and Fujii, Yasuhisa and Ratner, Alexander and Krishna, Ranjay and Lee, Chen-Yu and Pfister, Tomas},
  journal={arXiv preprint arXiv:2305.02301},
  year={2023}
}

@article{kim2021comparing,
  title={Comparing kullback-leibler divergence and mean squared error loss in knowledge distillation},
  author={Kim, Taehyeon and Oh, Jaehoon and Kim, NakYil and Cho, Sangwook and Yun, Se-Young},
  journal={arXiv preprint arXiv:2105.08919},
  year={2021}
}

@article{zhou2021bert,
  title={BERT learns to teach: Knowledge distillation with meta learning},
  author={Zhou, Wangchunshu and Xu, Canwen and McAuley, Julian},
  journal={arXiv preprint arXiv:2106.04570},
  year={2021}
}

@article{huang2022knowledge,
  title={Knowledge distillation from a stronger teacher},
  author={Huang, Tao and You, Shan and Wang, Fei and Qian, Chen and Xu, Chang},
  journal={Advances in Neural Information Processing Systems},
  volume={35},
  pages={33716--33727},
  year={2022}
}

@inproceedings{chen2021wasserstein,
  title={Wasserstein contrastive representation distillation},
  author={Chen, Liqun and Wang, Dong and Gan, Zhe and Liu, Jingjing and Henao, Ricardo and Carin, Lawrence},
  booktitle={Proceedings of the IEEE/CVF conference on computer vision and pattern recognition},
  pages={16296--16305},
  year={2021}
}

@article{lin2020ensemble,
  title={Ensemble distillation for robust model fusion in federated learning},
  author={Lin, Tao and Kong, Lingjing and Stich, Sebastian U and Jaggi, Martin},
  journal={Advances in neural information processing systems},
  volume={33},
  pages={2351--2363},
  year={2020}
}

@inproceedings{saha2022distilling,
  title={Distilling representational similarity using centered kernel alignment (cka)},
  author={Saha, Aninda and Bialkowski, Alina and Khalifa, Sara},
  booktitle={Proceedings of the the 33rd British Machine Vision Conference (BMVC 2022)},
  year={2022},
  organization={British Machine Vision Association}
}

@article{dapello2022aligning,
  title={Aligning model and macaque inferior temporal cortex representations improves model-to-human behavioral alignment and adversarial robustness},
  author={Dapello, Joel and Kar, Kohitij and Schrimpf, Martin and Geary, Robert and Ferguson, Michael and Cox, David D and DiCarlo, James J},
  journal={bioRxiv},
  pages={2022--07},
  year={2022},
  publisher={Cold Spring Harbor Laboratory}
}

@article{raghu2017svcca,
  title={Svcca: Singular vector canonical correlation analysis for deep learning dynamics and interpretability},
  author={Raghu, Maithra and Gilmer, Justin and Yosinski, Jason and Sohl-Dickstein, Jascha},
  journal={Advances in neural information processing systems},
  volume={30},
  year={2017}
}

@article{radford2019language,
  title={Language models are unsupervised multitask learners},
  author={Radford, Alec and Wu, Jeffrey and Child, Rewon and Luan, David and Amodei, Dario and Sutskever, Ilya and others},
  journal={OpenAI blog},
  volume={1},
  number={8},
  pages={9},
  year={2019}
}

@article{loshchilov2017decoupled,
  title={Decoupled weight decay regularization},
  author={Loshchilov, I},
  journal={arXiv preprint arXiv:1711.05101},
  year={2017}
}

@article{romero2014fitnets,
  title={Fitnets: Hints for thin deep nets},
  author={Romero, Adriana and Ballas, Nicolas and Kahou, Samira Ebrahimi and Chassang, Antoine and Gatta, Carlo and Bengio, Yoshua},
  journal={arXiv preprint arXiv:1412.6550},
  year={2014}
}

@inproceedings{subramaniam2025training,
  title={{Training the Untrainable: Introducing Inductive Bias via Representational Alignment}},
  author={Subramaniam, Vighnesh and Mayo, David and Conwell, Colin and Poggio, Tomaso and Katz, Boris and Cheung, Brian and Barbu, Andrei},
  booktitle={Advances in Neural Information Processing Systems (NeurIPS)},
  year={2025}
}

@article{zhang2024lolcats,
  title={Lolcats: On low-rank linearizing of large language models},
  author={Zhang, Michael and Arora, Simran and Chalamala, Rahul and Wu, Alan and Spector, Benjamin and Singhal, Aaryan and Ramesh, Krithik and R{\'e}, Christopher},
  journal={arXiv preprint arXiv:2410.10254},
  year={2024}
}

@article{mercat2024linearizing,
  title={Linearizing large language models},
  author={Mercat, Jean and Vasiljevic, Igor and Keh, Sedrick and Arora, Kushal and Dave, Achal and Gaidon, Adrien and Kollar, Thomas},
  journal={arXiv preprint arXiv:2405.06640},
  year={2024}
}

@article{zhang2024hedgehog,
  title={The hedgehog \& the porcupine: Expressive linear attentions with softmax mimicry},
  author={Zhang, Michael and Bhatia, Kush and Kumbong, Hermann and R{\'e}, Christopher},
  journal={arXiv preprint arXiv:2402.04347},
  year={2024}
}

@article{wang2024mamba,
  title={The mamba in the llama: Distilling and accelerating hybrid models},
  author={Wang, Junxiong and Paliotta, Daniele and May, Avner and Rush, Alexander and Dao, Tri},
  journal={Advances in Neural Information Processing Systems},
  volume={37},
  pages={62432--62457},
  year={2024}
}

@article{bick2024transformers,
  title={Transformers to ssms: Distilling quadratic knowledge to subquadratic models},
  author={Bick, Aviv and Li, Kevin and Xing, Eric and Kolter, J Zico and Gu, Albert},
  journal={Advances in Neural Information Processing Systems},
  volume={37},
  pages={31788--31812},
  year={2024}
}

@article{bick2025llamba,
  title={Llamba: Scaling distilled recurrent models for efficient language processing},
  author={Bick, Aviv and Katsch, Tobias and Sohoni, Nimit and Desai, Arjun and Gu, Albert},
  journal={arXiv preprint arXiv:2502.14458},
  year={2025}
}

@article{huh2024platonic,
  title={The platonic representation hypothesis},
  author={Huh, Minyoung and Cheung, Brian and Wang, Tongzhou and Isola, Phillip},
  journal={arXiv preprint arXiv:2405.07987},
  year={2024}
}

@article{zoph2016neural,
  title={Neural architecture search with reinforcement learning},
  author={Zoph, Barret and Le, Quoc V},
  journal={arXiv preprint arXiv:1611.01578},
  year={2016}
}

@inproceedings{real2019regularized,
  title={Regularized evolution for image classifier architecture search},
  author={Real, Esteban and Aggarwal, Alok and Huang, Yanping and Le, Quoc V},
  booktitle={Proceedings of the aaai conference on artificial intelligence},
  volume={33},
  number={01},
  pages={4780--4789},
  year={2019}
}

@article{liu2018darts,
  title={Darts: Differentiable architecture search},
  author={Liu, Hanxiao and Simonyan, Karen and Yang, Yiming},
  journal={arXiv preprint arXiv:1806.09055},
  year={2018}
}

@article{li2015convergent,
  title={Convergent learning: Do different neural networks learn the same representations?},
  author={Li, Yixuan and Yosinski, Jason and Clune, Jeff and Lipson, Hod and Hopcroft, John},
  journal={arXiv preprint arXiv:1511.07543},
  year={2015}
}

@article{han2015deep,
  title={Deep compression: Compressing deep neural networks with pruning, trained quantization and huffman coding},
  author={Han, Song and Mao, Huizi and Dally, William J},
  journal={arXiv preprint arXiv:1510.00149},
  year={2015}
}

@inproceedings{jacob2018quantization,
  title={Quantization and training of neural networks for efficient integer-arithmetic-only inference},
  author={Jacob, Benoit and Kligys, Skirmantas and Chen, Bo and Zhu, Menglong and Tang, Matthew and Howard, Andrew and Adam, Hartwig and Kalenichenko, Dmitry},
  booktitle={Proceedings of the IEEE conference on computer vision and pattern recognition},
  pages={2704--2713},
  year={2018}
}

@article{cai2019once,
  title={Once-for-all: Train one network and specialize it for efficient deployment},
  author={Cai, Han and Gan, Chuang and Wang, Tianzhe and Zhang, Zhekai and Han, Song},
  journal={arXiv preprint arXiv:1908.09791},
  year={2019}
}

@article{yu2018slimmable,
  title={Slimmable neural networks},
  author={Yu, Jiahui and Yang, Linjie and Xu, Ning and Yang, Jianchao and Huang, Thomas},
  journal={arXiv preprint arXiv:1812.08928},
  year={2018}
}

@article{oquab2023dinov2,
  title={Dinov2: Learning robust visual features without supervision},
  author={Oquab, Maxime and Darcet, Timoth{\'e}e and Moutakanni, Th{\'e}o and Vo, Huy and Szafraniec, Marc and Khalidov, Vasil and Fernandez, Pierre and Haziza, Daniel and Massa, Francisco and El-Nouby, Alaaeldin and others},
  journal={arXiv preprint arXiv:2304.07193},
  year={2023}
}

@article{loshchilov2016sgdr,
  title={Sgdr: Stochastic gradient descent with warm restarts},
  author={Loshchilov, Ilya and Hutter, Frank},
  journal={arXiv preprint arXiv:1608.03983},
  year={2016}
}

@article{alshammari2025unifying,
  title={I-con: A unifying framework for representation learning},
  author={Alshammari, Shaden and Hershey, John and Feldmann, Axel and Freeman, William T and Hamilton, Mark},
  journal={arXiv preprint arXiv:2504.16929},
  year={2025}
}

@inproceedings{ulyanov2018deep,
  title={Deep image prior},
  author={Ulyanov, Dmitry and Vedaldi, Andrea and Lempitsky, Victor},
  booktitle={Proceedings of the IEEE conference on computer vision and pattern recognition},
  pages={9446--9454},
  year={2018}
}

@article{bansal2021revisiting,
  title={Revisiting model stitching to compare neural representations},
  author={Bansal, Yamini and Nakkiran, Preetum and Barak, Boaz},
  journal={Advances in neural information processing systems},
  volume={34},
  pages={225--236},
  year={2021}
}

@article{conwell2024large,
  title={A large-scale examination of inductive biases shaping high-level visual representation in brains and machines},
  author={Conwell, Colin and Prince, Jacob S and Kay, Kendrick N and Alvarez, George A and Konkle, Talia},
  journal={Nature communications},
  volume={15},
  number={1},
  pages={9383},
  year={2024},
  publisher={Nature Publishing Group UK London}
}

@article{yu2024representation,
  title={Representation alignment for generation: Training diffusion transformers is easier than you think},
  author={Yu, Sihyun and Kwak, Sangkyung and Jang, Huiwon and Jeong, Jongheon and Huang, Jonathan and Shin, Jinwoo and Xie, Saining},
  journal={arXiv preprint arXiv:2410.06940},
  year={2024}
}

@book{plutarch_parallel_lives,
  author       = {Plutarch},
  title        = {Parallel Lives},
  origtitle    = {Βίοι Παράλληλοι},
  year         = {100--125},
  note         = {Composed ca. 100–125 CE.},
  location     = {Roman Empire},
  langid       = {english},
  origlanguage = {greek}
}

@article{zhong2024algorithmic,
  title={Algorithmic capabilities of random transformers},
  author={Zhong, Ziqian and Andreas, Jacob},
  journal={Advances in Neural Information Processing Systems},
  volume={37},
  pages={104357--104382},
  year={2024}
}

@article{geirhos2020surprising,
  title={On the surprising similarities between supervised and self-supervised models},
  author={Geirhos, Robert and Narayanappa, Kantharaju and Mitzkus, Benjamin and Bethge, Matthias and Wichmann, Felix A and Brendel, Wieland},
  journal={arXiv preprint arXiv:2010.08377},
  year={2020}
}

@inproceedings{zhang2024you,
  title={You only need less attention at each stage in vision transformers},
  author={Zhang, Shuoxi and Liu, Hanpeng and Lin, Stephen and He, Kun},
  booktitle={Proceedings of the IEEE/CVF Conference on Computer Vision and Pattern Recognition},
  pages={6057--6066},
  year={2024}
}

@article{venkataramanan2023skip,
  title={Skip-attention: Improving vision transformers by paying less attention},
  author={Venkataramanan, Shashanka and Ghodrati, Amir and Asano, Yuki M and Porikli, Fatih and Habibian, Amirhossein},
  journal={arXiv preprint arXiv:2301.02240},
  year={2023}
}

@article{yu2023metaformer,
  title={Metaformer baselines for vision},
  author={Yu, Weihao and Si, Chenyang and Zhou, Pan and Luo, Mi and Zhou, Yichen and Feng, Jiashi and Yan, Shuicheng and Wang, Xinchao},
  journal={IEEE Transactions on Pattern Analysis and Machine Intelligence},
  volume={46},
  number={2},
  pages={896--912},
  year={2023},
  publisher={IEEE}
}

@inproceedings{wang2023riformer,
  title={Riformer: Keep your vision backbone effective but removing token mixer},
  author={Wang, Jiahao and Zhang, Songyang and Liu, Yong and Wu, Taiqiang and Yang, Yujiu and Liu, Xihui and Chen, Kai and Luo, Ping and Lin, Dahua},
  booktitle={Proceedings of the IEEE/CVF conference on computer vision and pattern recognition},
  pages={14443--14452},
  year={2023}
}

@article{bachmann2023scaling,
  title={Scaling mlps: A tale of inductive bias},
  author={Bachmann, Gregor and Anagnostidis, Sotiris and Hofmann, Thomas},
  journal={Advances in Neural Information Processing Systems},
  volume={36},
  pages={60821--60840},
  year={2023}
}

@article{tolstikhin2021mlp,
  title={Mlp-mixer: An all-mlp architecture for vision},
  author={Tolstikhin, Ilya O and Houlsby, Neil and Kolesnikov, Alexander and Beyer, Lucas and Zhai, Xiaohua and Unterthiner, Thomas and Yung, Jessica and Steiner, Andreas and Keysers, Daniel and Uszkoreit, Jakob and others},
  journal={Advances in neural information processing systems},
  volume={34},
  pages={24261--24272},
  year={2021}
}

@inproceedings{song2007supervised,
  title={Supervised feature selection via dependence estimation},
  author={Song, Le and Smola, Alex and Gretton, Arthur and Borgwardt, Karsten M and Bedo, Justin},
  booktitle={Proceedings of the 24th international conference on Machine learning},
  pages={823--830},
  year={2007}
}

@inproceedings{xu-etal-2020-bert,
    title = "{BERT}-of-Theseus: Compressing {BERT} by Progressive Module Replacing",
    author = "Xu, Canwen  and
      Zhou, Wangchunshu  and
      Ge, Tao  and
      Wei, Furu  and
      Zhou, Ming",
    booktitle = "Proceedings of the 2020 Conference on Empirical Methods in Natural Language Processing (EMNLP)",
    month = nov,
    year = "2020",
    address = "Online",
    publisher = "Association for Computational Linguistics",
    url = "https://www.aclweb.org/anthology/2020.emnlp-main.633",
    pages = "7859--7869"
}

@misc{chen2023vanillanetpowerminimalismdeep,
      title={VanillaNet: the Power of Minimalism in Deep Learning}, 
      author={Hanting Chen and Yunhe Wang and Jianyuan Guo and Dacheng Tao},
      year={2023},
      eprint={2305.12972},
      archivePrefix={arXiv},
      primaryClass={cs.CV},
      url={https://arxiv.org/abs/2305.12972}, 
}

@misc{pytorch_reducelronplateau,
  title        = {{ReduceLROnPlateau --- PyTorch 2.9 documentation}},
  author       = {{PyTorch Contributors}},
  howpublished = {\url{https://docs.pytorch.org/docs/stable/generated/torch.optim.lr_scheduler.ReduceLROnPlateau.html}},
  note         = {Accessed: 2025-11-28},
  year         = {2025}
}

@inproceedings{he2022feature,
  title={Feature kernel distillation},
  author={He, Bobby and Ozay, Mete},
  booktitle={International conference on learning representations},
  year={2022}
}

%%%%%%%%%%%%%%%%%%%%%%%%%%%%%%%%%%%%%%%%%%%%%%%%%%%%%%%%%%%%
\newpage
\appendix

\section{Appendix Overview}
\begin{displayquote}
The ship wherein Theseus and the youth of Athens returned had thirty oars, and was preserved by the Athenians down even to the time of Demetrius Phalereus, for they took away the old planks as they decayed, putting in new and stronger timber in their place, insomuch that this ship became a standing example among the philosophers, for the logical question of things that grow; one side holding that the ship remained the same, and the other contending that it was not the same.

--Plutarch
\end{displayquote}

We present additional details of NoT, experiments and analysis, as well as additional results. In \Cref{ap:methods}, we provide further details of NoT, providing an overview of the method and covering the similarity metrics applied in this paper, CKA and D-MNN in detail. In \Cref{ap:exp}, we provide additional details on our experiments, going into further detail on training and evaluation. We cover our batch normalization recalibration as well as general details on our ResNet-50$\rightarrow$ResNet-18 result. In \Cref{ap:dmnn_staged}, we recreate \Cref{fig:align_loss} with D-MNN for further analysis with D-MNN. In \Cref{ap:hidden_adapters}, we compare hidden-state adapters and logit distillation. In \Cref{ap:efficiency}, we report inference latency measurements. In \Cref{ap:dino}, we provide further analysis of DINOv2 to investigate why we can remove cross-token calculations. 

% In \Cref{ap:loss}, we provide visualizations of our loss curves during similarity optimization with CKA to show our current training dynamics with NoT across representational similarity training.

\section{Methods Overview}
\label{ap:methods}
We give an overview of NoT in \Cref{alg:not} and cover more methodological details here. 
\begin{algorithm}
\caption{\textbf{Network of Theseus (NoT):} staged representational alignment and conversion}
\label{alg:not}
\begin{algorithmic}[1]
\small
\Require Guide $f^G(\cdot;\theta^G)$ (frozen), target $f^T$ (init as $f^G$), replacement mapping $\mathcal R$ (shape-compatible), schedule $\mathcal S=(I_1,\dots,I_T)$ (default progressive), unlabeled inputs $\mathcal X$, labeled data $\mathcal D=\{(x,y)\}$, similarity $g$ (e.g., CKA); define $\Delta(\mA,\mB)=1-g(\mA,\mB)$
\For{$t=1$ to $T$} \Comment{Alignment stage $t$}
  \State \textcolor{gray}{\# Replace layer $t$, assume previous layers have been replaced}
  \State \colorbox{superlightred}{$J_t \gets I_t \setminus I_{t-1}$; \textbf{replace} $\{i \in J_t\}$ in $f^T$ with $\{\mathcal R(i)\}$ (initialize)}
  \State \textbf{freeze} $\theta^T_{\overline{I_t}}$ and \textbf{unfreeze} $\theta^T_{I_t}$ \Comment{no drift outside replaced set}
  \For{mini-batches $x \sim \mathcal X$}
    \State For each $i \in I_t$: $\mA_i^G \leftarrow$ activations of layer $i$ in $f^G(x)$; $\mA_i^T \leftarrow$ activations of layer $i$ in $f^T(x)$
    \State \colorbox{lightblue}{$L_{\text{align}} \leftarrow \Delta(\mA^T_i, \mA_i^G)$}
    \State Update $\theta^T_{I_t}$ to minimize $L_{\text{align}}$
  \EndFor
\EndFor
\State \textbf{Fine-tune}: unfreeze all $\theta^T$ and minimize $L_{\text{task}}(f^T(x),y)$ over $(x,y)\in \mathcal D$
\State \Return converted target $f^T$
\end{algorithmic}
\end{algorithm}

\subsection{Centered Kernel Alignment}
\label{ap:cka}
To compare representations, we use a representation similarity metric, $\mathcal{M}$, which corresponds to centered kernel alignment (CKA) \citep{kornblith2019similarity, cortes2012algorithms, cristianini2001kernel} in our setting. We specifically consider linear CKA. 

CKA uses kernel functions on mean-centered representations to compute representational similarity matrices, which are then compared via the Hilbert-Schmidt Independence Criterion (HSIC). More specifically, suppose we have two sets of representations $\mA \in \mathbb{R}^{b\times d_1}$ and $\mB \in \mathbb{R}^{b \times d_2}$. We first compute the Gram matrices for each set of representations

\begin{equation}
    \label{eq:gram}
    \mK = \mA\mA^T, \mL = \mB'\mB'^T
\end{equation}

We center the Gram matrices by introducing a matrix, $H$, where $H = \mI_b - \frac{1}{b}\vone\vone^T$. 

\begin{equation}
    \tilde{\mK} = \mH\mK\mH, \tilde{\mL} = \mH\mL\mH
\end{equation}

We compute the HSIC on the Gram matrices.

\begin{equation}
    HSIC(\mK, \mL) = \text{tr}(\tilde{\mK}, \tilde{\mL})
\end{equation}

Finally, we define our linear CKA metric as:
\begin{equation}
    g_{\text{CKA}}(\mA, \mB'):= \text{CKA}(\mK, \mL) = \frac{HSIC(\mK, \mL)}{\sqrt{HSIC(\mK, \mK) * HSIC(\mL, \mL)}}
\end{equation}

In our setting, we consider representational \emph{dissimilarity} and aim to minimize the dissimilarity between representations from our target network and guide network. We define this as:

\begin{equation}
    \Delta_{\text{CKA}}(\mA, \mB) = 1 - g(\mA, \mB)
\end{equation}

Linear CKA ranges from $0$ to $1$ (very different representations). Because of this, we take the complement by subtracting the linear CKA from $1$ to represent dissimilarity. Most importantly, before computing $g$, rows of $\mA$ and $\mB$ are $\ell_2$-normalized.

\subsubsection{Unbiased CKA}
\label{ap:ucka}
We use an unbiased estimator for HSIC for our GPT-2 Large$\rightarrow$GPT-2 Small experiments, motivated by findings covered in \Cref{ap:gpt2_ucka}. We cover the unbiased HSIC here. 

We start with our Gram matrices, $\mK, \mL$. An unbiased estimator for HSIC then involves forming zero-diagonal copies of both matrices. 

\begin{equation}
\tilde \mK \coloneqq \mK - \operatorname{diag}(\mK), \qquad 
\tilde \mL \coloneqq \mL - \operatorname{diag}(\mL),
\end{equation}

so that $\tilde \mK_{ii}=\tilde \mL_{ii}=0$ and $\tilde \mK_{ij}=\mK_{ij}$, $\tilde \mL_{ij}=L_{ij}$ for $i\neq j$. Let $\vone\in\mathbb{R}^m$ be the all-ones vector. The unbiased estimator of HSIC is

\begin{equation}
\label{eq:hsic-unbiased}
HSIC_{U}(K,L)
= \frac{1}{m(m-3)}
\left[
\operatorname{tr}(\tilde \mK \tilde \mL)
+ \frac{\operatorname{tr}(\tilde \mK\tilde \mL)}{(m-1)(m-2)}
- \frac{2}{m-2}\,\operatorname{tr}(\tilde \mK\tilde \mL)
\right],
\end{equation}
which is well-defined for $m\ge 4$. The last term corresponds to the unbiased correction. 

\subsection{Differentiable Mutual Nearest Neighbors}
\label{ap:dknn}
For Differentiable MNN (D-MNN), we use \citet{huh2024platonic} and \citet{alshammari2025unifying} to reformulate nearest neighbors into a differentiable form.

Let $b$ be the mini-batch size and anchors $i\in\{1,\dots,b\}$.
Two encoders $f_1, f_2$ produce features $\phi_i=f_1(x_i)\in\mathbb{R}^d$,
$\psi_i=f_2(y_i)\in\mathbb{R}^d$.
Define $\mathcal{J}_i=\{1,\dots,b\}\setminus\{i\}$.
For $k\in\mathbb{N}$, let $s_{f_1}(i),s_{f_2}(i)\subset\mathcal{J}_i$ be the index sets of the $k$ nearest neighbors of $i$ under $\{\phi_j\}$ and, $\{\psi_j\}$, respectively.

\textbf{k-Nearest Neighbors}: As defined by \citet{huh2024platonic}, the overlap score for anchor $i$ is
\begin{equation}
g_{NN}(\phi_i, \psi_i) = \frac{|s_{f_1}(i), s_{f_2}(i)|}{k}
\end{equation}
and the batch score is the average over $i$.

\textbf{From overlap to probabilities}:
We introduce hard conditional distributions uniform on the kNN sets:
\begin{equation}
  p_{f_1}(j\mid i)=\frac{\vone\{j\in s_{f_1}(i)\}}{k},\qquad
  q_{f_2}(j\mid i)=\frac{\vone\{j\in s_{f_2}(i)\}}{k},\quad j\in\mathcal{J}_i.
\end{equation}
Then
\begin{equation}
  \sum_{j\in\mathcal{J}_i} p_{f_1}(j\mid i)\,q_{f_2}(j\mid i)=\frac{|s_{f_1}(i)\cap s_{f_2}(i)|}{k^2},
  \quad\Rightarrow\quad
  g_{NN}(\phi_i,\psi_i)=k\sum_{j\in\mathcal{J}_i} p_{f_1}(j\mid i)\,q_{f_2}(j\mid i).
  \label{eq:nn-prob}
\end{equation}

\textbf{Differentiable conditional neighborhoods}: 
Let $\tau>0$ be a temperature. Define pairwise similarities either by
\begin{equation}
  \label{eq:temp}
  s^f_{ij}=-\frac{\|\phi_i-\phi_j\|_2^2}{\tau}
  \quad\text{or}\quad
  s^f_{ij}=\frac{\phi_i^\top\phi_j}{\tau},
  \qquad j\in\mathcal{J}_i,
\end{equation}
and analogously $s^g_{ij}$ from $\{\psi_j\}$.
Let $T_k^f(i)\subset\mathcal{J}_i$ and $T_k^g(i)\subset\mathcal{J}_i$ denote top-$k$ index selections under $s^f_{ij}$ and $s^g_{ij}$, respectively. We define masked, normalized conditionals over the top-$k$:
\begin{equation}
  P_{ij} \equiv p_{f_1}(j\mid i)
  = \frac{\exp(s^{f_1}_{ij})\,\vone\{j\in T_k^{f_1}(i)\}}
         {\sum_{j'\in T_k^{f_1}(i)} \exp(s^{f_1}_{ij'})},
  \qquad
  Q_{ij} \equiv q_{f_2}(j\mid i)
  = \frac{\exp(s^{f_2}_{ij})\,\vone\{j\in T_k^{f_2}(i)\}}
         {\sum_{j'\in T_k^{f_2}(i)} \exp(s^{f_2}_{ij'})}.
  \label{eq:masked-softmax}
\end{equation}
(Any differentiable soft-top-$k$ operator may be used; with an exact mask and $\tau\!\to\!0$,
$P_{ij}$ and $Q_{ij}$ converge to the hard uniforms.)

\textbf{D-MNN alignment and limit}:
We define the differentiable alignment for anchor $i$ as the inner product
\begin{equation}
  g_{\text{soft}}(\phi_i,\psi_i) \;=\; \sum_{j\in\mathcal{J}_i} P_{ij}\,Q_{ij},
  \qquad
  \bar g_{\text{soft}} \;=\; \frac{1}{b}\sum_{i=1}^b m_{\text{soft}}(\phi_i,\psi_i).
\end{equation}
Under the exact top-$k$ mask and $\tau\!\to\!0$,
$P_{ij},Q_{ij}\!\to\!\frac{1}{k}\vone\{j\in s_{f_1}(i)\},\frac{1}{k}\vone\{j\in s_{f_2}(i)\}$,
so that $k\,g_{\text{soft}}(\phi_i,\psi_i)\to g_{NN}(\phi_i,\psi_i)$ by \eqref{eq:nn-prob}.

\textbf{Training objective}:
Excluding the anchor, we write $P_{i\setminus *}=\{P_{ij}\}_{j\in\mathcal{J}_i}$ and similarly for $Q$.
When $P$ (from $f$) guides $Q$ (from $g$), we minimize
\begin{equation}
  \Delta_{\text{D-MNN}} \;:=\; \mathcal{L}_{KL} \;=\; \frac{1}{b}\sum_{i=1}^b
  D_{KL}\!\Big(P_{i\setminus *}\,\Big\|\,Q_{i\setminus *}\Big).
  \label{eq:dknn-kl}
\end{equation}
This KL aligns conditional neighborhood distributions. Most importantly, before computing $g$, rows of $\mA$ and $\mB$ are $\ell_2$-normalized. For our setting in this paper, we have to tune the temperature for D-MNN.

\subsection{Methodology Limitations}
NoT has a number of limitations as covered previously. We go deeper into methodological limitations here. NoT can be expensive memory-wise and runtime-wise. Each stage of NoT requires independent tuning. We also find that training can take a significant amount of time. For some conversions, we ran some conversions for up to 50 epochs of training for the best conversion. We believe these limitations may be overcome with further training optimizations and techniques such as more rich data augmentation and stronger label smoothing as used in previous papers \citep{bachmann2023scaling}. 

\section{Experimental Details}
% \usepackage{multirow}
\begin{table}[ht]
\begin{adjustbox}{max width=\textwidth}

\begin{tabular}{cccccc}
\midrule
\multicolumn{1}{l}{Model}                               & Stage & Trained Learning Rate & Untrained Learning Rate & D-MNN Learning Rate & Number of Epochs \\ \midrule
\multirow{20}{*}{ResNet-18$\rightarrow$MLP}             & 1     & $2\times 10^{-5}$     & $2\times 10^{-5}$       & $1\times 10^{-5}$   & 10               \\
                                                        & 2     & $2\times 10^{-5}$     & $2\times 10^{-5}$       & $1\times 10^{-5}$   & 65               \\
                                                        & 3     & $2\times 10^{-5}$     & $2\times 10^{-5}$       & $7.5\times 10^{-6}$ & 45               \\
                                                        & 4     & $1.5\times 10^{-5}$   & $1.5\times 10^{-5}$     & $7.5\times 10^{-6}$ & 45               \\
                                                        & 5     & $1.5\times 10^{-5}$   & $1.5\times 10^{-5}$     & $7.5\times 10^{-6}$ & 45               \\
                                                        & 6     & $1\times 10^{-5}$     & $1\times 10^{-5}$       & $7.5\times 10^{-6}$ & 65               \\
                                                        & 7     & $1\times 10^{-5}$     & $1\times 10^{-5}$       & $7.5\times 10^{-6}$ & 45               \\
                                                        & 8     & $1\times 10^{-5}$     & $1\times 10^{-5}$       & $5\times 10^{-6}$   & 45               \\
                                                        & 9     & $1\times 10^{-5}$     & $1\times 10^{-5}$       & $5\times 10^{-6}$   & 50               \\
                                                        & 10    & $7.5\times 10^{-6}$   & $7.5\times 10^{-6}$     & $5\times 10^{-6}$   & 50               \\
                                                        & 11    & $7.5\times 10^{-6}$   & $7.5\times 10^{-6}$     & $5\times 10^{-6}$   & 50               \\
                                                        & 12    & $7.5\times 10^{-6}$   & $7.5\times 10^{-6}$     & $5\times 10^{-6}$   & 50               \\
                                                        & 13    & $7.5\times 10^{-6}$   & $7.5\times 10^{-6}$     & $5\times 10^{-6}$   & 65               \\
                                                        & 14    & $7.5\times 10^{-6}$   & $7.5\times 10^{-6}$     & $1\times 10^{-6}$   & 50               \\
                                                        & 15    & $5\times 10^{-6}$     & $5\times 10^{-6}$       & $1\times 10^{-6}$   & 50               \\
                                                        & 16    & $5\times 10^{-6}$     & $5\times 10^{-6}$       & $1\times 10^{-6}$   & 50               \\
                                                        & 17    & $2.5\times 10^{-6}$   & $2.5\times 10^{-6}$     & $1\times 10^{-6}$   & 75               \\
                                                        & 18    & $2.5\times 10^{-6}$   & $2.5\times 10^{-6}$     & $1\times 10^{-6}$   & 75               \\
                                                        & 19    & $1\times 10^{-6}$     & $1\times 10^{-6}$       & $1\times 10^{-6}$   & 75               \\
                                                        & 20    & $1\times 10^{-6}$     & $1\times 10^{-6}$       & $1\times 10^{-6}$   & 100              \\ \midrule
\multirow{12}{*}{DINOv2$\rightarrow$Patch-MLP}          & 1     & $1\times 10^{-4}$     & $1\times 10^{-4}$       & $1\times 10^{-4}$   & 10               \\
                                                        & 2     & $1\times 10^{-4}$     & $1\times 10^{-4}$       & $1\times 10^{-4}$   & 35               \\
                                                        & 3     & $1\times 10^{-4}$     & $1\times 10^{-4}$       & $1\times 10^{-4}$   & 35               \\
                                                        & 4     & $1\times 10^{-4}$     & $1\times 10^{-4}$       & $7.5\times 10^{-5}$ & 35               \\
                                                        & 5     & $1\times 10^{-4}$     & $1\times 10^{-4}$       & $7.5\times 10^{-5}$ & 50               \\
                                                        & 6     & $1\times 10^{-4}$     & $1\times 10^{-4}$       & $7.5\times 10^{-5}$ & 50               \\
                                                        & 7     & $1\times 10^{-4}$     & $1\times 10^{-4}$       & $5\times 10^{-5}$   & 50               \\
                                                        & 8     & $1\times 10^{-4}$     & $1\times 10^{-4}$       & $5\times 10^{-5}$   & 50               \\
                                                        & 9     & $7.5\times 10^{-5}$   & $7.5\times 10^{-5}$     & $5\times 10^{-5}$   & 60               \\
                                                        & 10    & $7.5\times 10^{-5}$   & $7.5\times 10^{-5}$     & $5\times 10^{-5}$   & 75               \\
                                                        & 11    & $7.5\times 10^{-5}$   & $7.5\times 10^{-5}$     & $1\times 10^{-5}$   & 75               \\
                                                        & 12    & $7.5\times 10^{-5}$   & $7.5\times 10^{-5}$     & $1\times 10^{-5}$   & 100              \\ \midrule
\multirow{12}{*}{GPT-2$\rightarrow$RNN}                 & 1     & $1\times 10^{-4}$     & $1\times 10^{-4}$       & $1\times 10^{-4}$   & 10               \\
                                                        & 2     & $1\times 10^{-4}$     & $1\times 10^{-4}$       & $1\times 10^{-4}$   & 35               \\
                                                        & 3     & $1\times 10^{-4}$     & $1\times 10^{-4}$       & $1\times 10^{-4}$   & 35               \\
                                                        & 4     & $1\times 10^{-4}$     & $1\times 10^{-4}$       & $1\times 10^{-4}$   & 50               \\
                                                        & 5     & $1\times 10^{-4}$     & $1\times 10^{-4}$       & $1\times 10^{-4}$   & 50               \\
                                                        & 6     & $1\times 10^{-4}$     & $1\times 10^{-4}$       & $1\times 10^{-4}$   & 50               \\
                                                        & 7     & $1\times 10^{-4}$     & $1\times 10^{-4}$       & $1\times 10^{-4}$   & 50               \\
                                                        & 8     & $1\times 10^{-4}$     & $1\times 10^{-4}$       & $1\times 10^{-4}$   & 50               \\
                                                        & 9     & $1\times 10^{-4}$     & $1\times 10^{-4}$       & $1\times 10^{-4}$   & 50               \\
                                                        & 10    & $1\times 10^{-4}$     & $1\times 10^{-4}$       & $1\times 10^{-4}$   & 50               \\
                                                        & 11    & $1\times 10^{-4}$     & $1\times 10^{-4}$       & $1\times 10^{-4}$   & 50               \\
                                                        & 12    & $1\times 10^{-4}$     & $1\times 10^{-4}$       & $1\times 10^{-4}$   & 50               \\ \midrule
\multirow{8}{*}{ResNet-50$\rightarrow$ResNet-18}        & 1     & $1\times 10^{-3}$     & $1\times 10^{-3}$       & $1\times 10^{-3}$   & 5                \\
                                                        & 2     & $1\times 10^{-3}$     & $1\times 10^{-3}$       & $1\times 10^{-3}$   & 15               \\
                                                        & 3     & $1\times 10^{-3}$     & $1\times 10^{-3}$       & $1\times 10^{-3}$   & 25               \\
                                                        & 4     & $1\times 10^{-3}$     & $1\times 10^{-3}$       & $1\times 10^{-3}$   & 35               \\
                                                        & 5     & $1\times 10^{-3}$     & $1\times 10^{-3}$       & $1\times 10^{-3}$   & 35               \\
                                                        & 6     & $1\times 10^{-3}$     & $1\times 10^{-3}$       & $1\times 10^{-3}$   & 35               \\
                                                        & 7     & $1\times 10^{-3}$     & $1\times 10^{-3}$       & $1\times 10^{-3}$   & 75               \\
                                                        & 8     & $1\times 10^{-3}$     & $1\times 10^{-3}$       & $1\times 10^{-3}$   & 100              \\ \midrule
\multirow{12}{*}{GPT-2 Large $\rightarrow$ GPT-2 Small} & 1     & $7.5\times 10^{-5}$   & $7.5\times 10^{-5}$     & $1\times 10^{-5}$   & 15               \\
                                                        & 2     & $7.5\times 10^{-5}$   & $7.5\times 10^{-5}$     & $1\times 10^{-5}$   & 15               \\
                                                        & 3     & $7.5\times 10^{-5}$   & $7.5\times 10^{-5}$     & $1\times 10^{-5}$   & 15               \\
                                                        & 4     & $7.5\times 10^{-5}$   & $7.5\times 10^{-5}$     & $1\times 10^{-5}$   & 15               \\
                                                        & 5     & $7.5\times 10^{-5}$   & $7.5\times 10^{-5}$     & $1\times 10^{-5}$   & 15               \\
                                                        & 6     & $7.5\times 10^{-5}$   & $7.5\times 10^{-5}$     & $1\times 10^{-5}$   & 15               \\
                                                        & 7     & $7.5\times 10^{-5}$   & $7.5\times 10^{-5}$     & $1\times 10^{-5}$   & 15               \\
                                                        & 8     & $7.5\times 10^{-5}$   & $7.5\times 10^{-5}$     & $1\times 10^{-5}$   & 15               \\
                                                        & 9     & $7.5\times 10^{-5}$   & $7.5\times 10^{-5}$     & $1\times 10^{-5}$   & 15               \\
                                                        & 10    & $7.5\times 10^{-5}$   & $7.5\times 10^{-5}$     & $1\times 10^{-5}$   & 15               \\
                                                        & 11    & $7.5\times 10^{-5}$   & $7.5\times 10^{-5}$     & $1\times 10^{-5}$   & 15               \\
                                                        & 12    & $7.5\times 10^{-5}$   & $7.5\times 10^{-5}$     & $1\times 10^{-5}$   & 15               \\ \bottomrule
\end{tabular}
\end{adjustbox}
\caption{\textbf{Learning rates and epochs of training across NoT representational similarity optimization}: We report learning rates and number of epochs of training used for training our model replacements for NoT representational similarity matching. We find that different stages can have significantly different learning rate requirements.}
\vspace{-2ex}
\label{tab:lrs_epochs}
\end{table}

\begin{table}[ht]
\begin{adjustbox}{max width=\textwidth}    
\begin{tabular}{cccccc}\toprule
Model                           & Learning Rate     & Epochs & Warm-up & Scheduler     & Notes                    \\ \toprule
ResNet-18$\rightarrow$MLP       & $5\times 10^{-4}$ & 100    & 10      & Linear+Cosine & Gradient Clipping to 1.0 \\
DINOv2$\rightarrow$Patch-MLP    & $1\times 10^{-4}$ & 100    & 5       & Linear+Cosine & N/A                      \\
ResNet-50$\rightarrow$ResNet-18 & $1\times 10^{-3}$ & 100    & 0       & None          & N/A                      \\
GPT-2$\rightarrow$RNN           & $1\times 10^{-4}$ & 100    & 5       & Linear+Cosine & Gradient Clipping to 1.0 \\ \bottomrule
\end{tabular}
\end{adjustbox}
\caption{\textbf{Learning rates, epochs, and scheduler settings for NoT task training.} We report learning rates, number of epochs of training, number of warmup epochs, and scheduler details for our settings along with miscellaneous details like using gradient clipping.}
\label{tab:task}
\end{table}
\label{ap:exp}

All training experiments were done on 8 H100 GPUs over 2 months in total. GPU optimization techniques were taken such as gradient accumulation and gradient
checkpointing and some experiments with converting ResNet-18 to an MLP used mixed-precision training.

In \Cref{tab:lrs_epochs}, we show a table of learning rates used at each stage of training for our replacements. We also show our task training details in \Cref{tab:task}. 

\subsection{Handling BatchNorm}
A difficulty with NoT is handling networks which use batch normalization, a common design choice in convolutional networks. However, in NoT, we find that batch normalization requires additional design considerations to ensure proper architectural transfer. For example, batch-feature statistics may shift for the replaced layers, meaning that batch normalization may need changes. We find that recalibrating batch normalization during representational similarity optimization has adverse effects on downstream performance. 

To mitigate this, we freeze batch normalization calculations in the guide network during replacement. This also makes the replacement more difficult for the target modules. After representational similarity optimization, we recalibrate the batch normalization layers in the resultant target architecture before task training. The recalibration only focuses on tuning the batch normalization layers before doing target training. We find that this separation is useful, changing performance by a significant amount. 

\subsection{ResNet-50 Replacement}
\label{ap:rn50}
Beyond direct layer/block replacements, we generalize NoT across architectures to convert from one architecture to another more generally. To do this, we relax the constraint requiring replaced components of our guide network to functionally fit into the guide network i.e. the replaced target module must accept the correct input and return the correct output. Having this constraint was necessary for \Cref{fig:layer_replace} to train different hybrid networks as we replace layers in the guide network. 

See \Cref{fig:resnet18_theseus}. ResNet-50 blocks can be replaced with ResNet-18 blocks without incorporation into the full architecture. A full hybrid architecture with ResNet-18 and ResNet-50 blocks is unnecessary in this setting since the main goal is to reconstruct ResNet-18 while optimizing representational similarity with ResNet-50. Our findings shown in \Cref{tab:main} and \Cref{tab:untrained} are exciting because it demonstrates that we can make conversion across architectures general. For example, in \Cref{fig:resnet18_theseus}, ResNet-50 can be replaced with any architecture like GPT-2, and this can be converted to a smaller transformer or a smaller RNN. 

\section{Biased vs Unbiased CKA: GPT-2 Large\textrightarrow GPT-2 Small}
\label{ap:gpt2_ucka}
\begin{figure}
    \centering
    \includegraphics[width=0.6\textwidth]{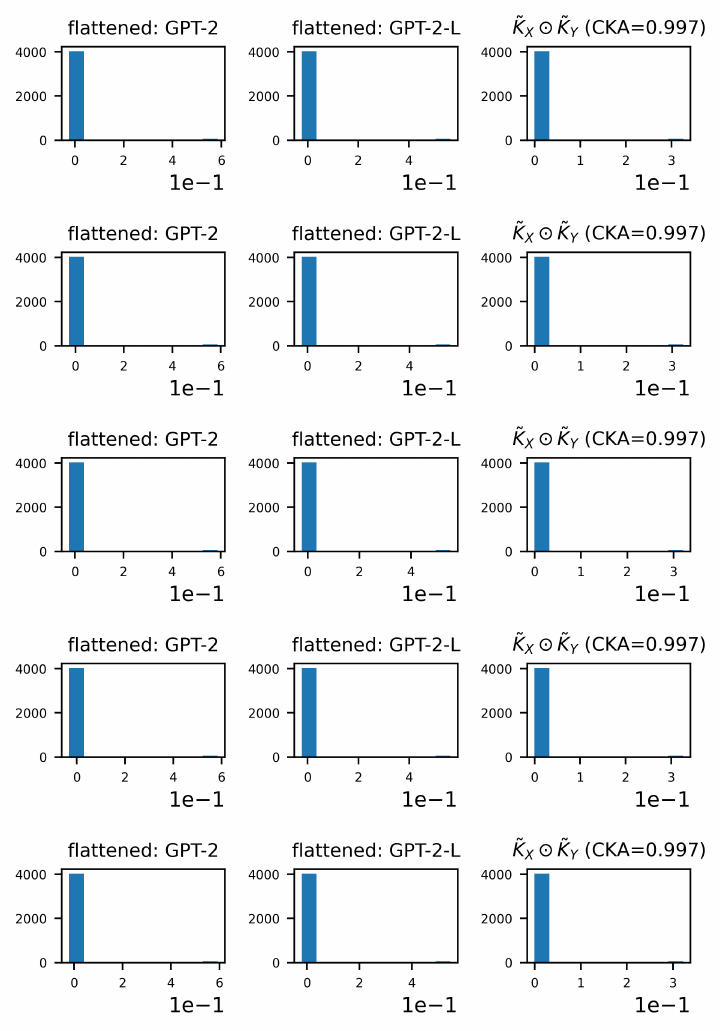}
    \caption{\textbf{Biased CKA inflated similarity scores between GPT-2 Small and GPT-2 Large.} When plotting values in the Gram matrices of a random pair of layers from GPT-2 Small and GPT-2 Large, we find that the Gram matrix values are very skewed. This is lead by inflated diagonal scores and skewed Gram matrix comparisons. We fix this using an unbiased HSIC estimator.}
    \label{fig:grams}
\end{figure}
In our GPT-2 Large$\rightarrow$GPT-2 Small setting, we found that using biased CKA as our alignment metric between GPT-2 Small transformer blocks and three GPT-2 Large transformer blocks did not have significant improvement. This is explained by inflated biased CKA values between GPT-2 Large blocks and GPT-2 Small blocks. We found that starting CKA values between randomly initialized GPT-2 Small blocks and trained GPT-2 Large blocks started at 0.98, leading to very small optimization gains from optimizing CKA and very small improvements on downstream performance with Wikitext-103. 

To better understand this trend, we plotted the distribution of values in our Gram matrices as a histogram in \Cref{fig:grams}. These plots correspond with visualizing the Gram matrices for 5 batches of 256 examples. The first two columns correspond to a Gram matrix as described by \Cref{eq:gram} and the third columns corresponds to the product across the Gram matrices of the two models used by HSIC. We can see that the Gram matrices are heavily skewed with some values appearing to be much higher. These values correspond with the self-similarity in the networks, leading to inflated CKA scores. To mitigate this, we removed computing the HSIC with the diagonal using unbiased CKA \citep{song2007supervised}, which we discussed in \Cref{ap:ucka}.

We applied UCKA in other experimental settings but found either degraded or comparable results to using CKA. 

\section{D-MNN Staged Alignment}
\label{ap:dmnn_staged}
\begin{figure}
    \centering
    \includegraphics[width=\textwidth]{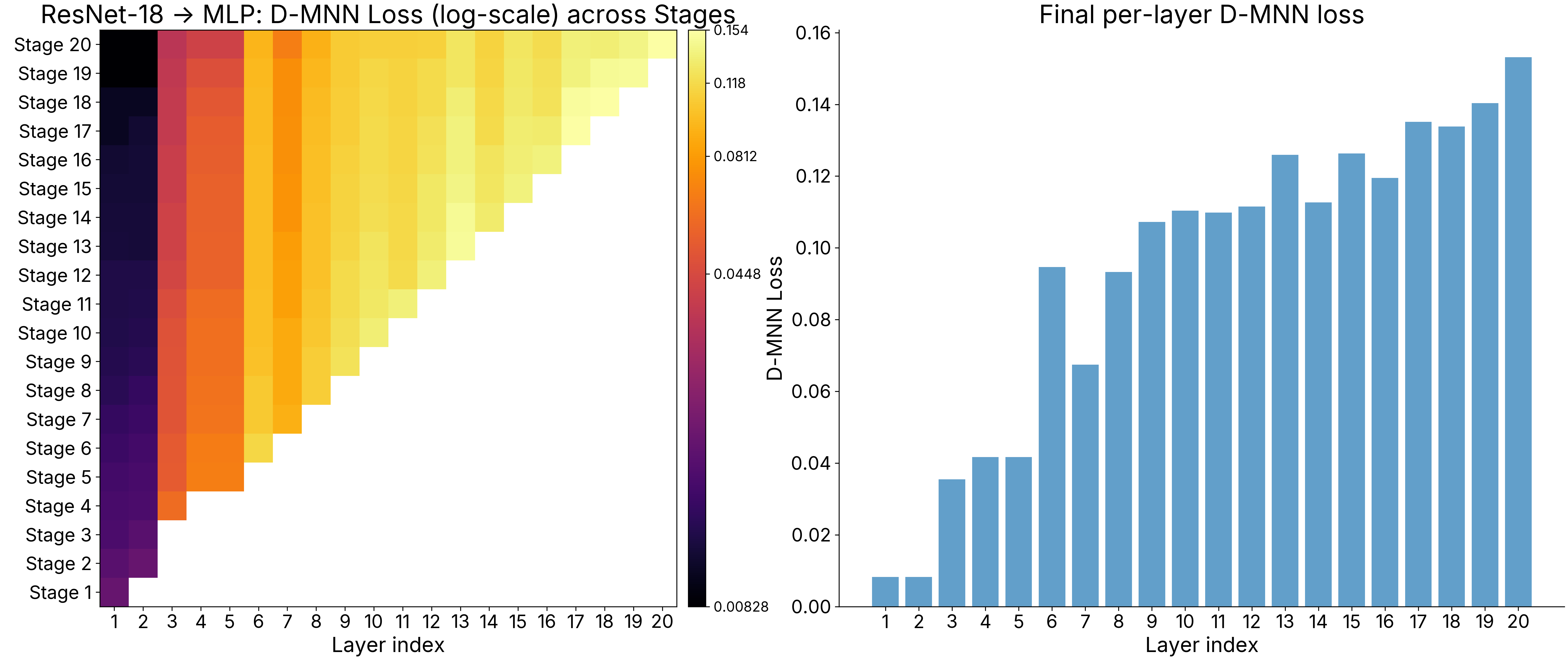}
    \caption{\textbf{D-MNN finds similar layer bottlenecks as CKA.} We show the per-stage losses for D-MNN, as with CKA. We find that D-MNN is far from saturating unlike CKA, which had loss values closer to 0. We find that D-MNN is also more consistent across layers, indicating that further optimization can be pushed for better alignment in D-MNN. We believe this shows even better correlation between D-MNN and final task accuracy and optimizing D-MNN further may lead to a task accuracy closer to the guide network.}
    \label{fig:dmnn_align_loss}
\end{figure}
We recreate \Cref{fig:align_loss} for our new metric, D-MNN (\Cref{ap:dknn}) to analyze loss dynamics. We show results in \Cref{fig:dmnn_align_loss}. We find that the quantitative loss for D-MNN is much higher and farther from 0, indicating little saturation. This is likely because D-MNN prioritizes locality over global geometry when comparing representations. We believe this demonstrates the benefit of our method.

\section{Reversing NoT Architecture Replacement}
\begin{table}[t]
\centering
\begin{tabular}{lcccc}
\toprule
Original (Guide) $\rightarrow$ Target & Similarity Metric & Guide & NoT & Baseline \\
\midrule
& & \multicolumn{3}{c}{\textbf{ImageNet Top-1 Accuracy} ($\uparrow$)} \\
\cmidrule(lr){3-5}
MLP (Untrained) $\rightarrow$ ResNet-18 & CKA  & 0.0911 & 68.06 & 68.90 \\
MLP (Trained) $\rightarrow$ ResNet-18 & CKA & 32.18 & 67.54 & 68.90 \\
RNN (Untrained) $\rightarrow$ ResNet-18 & CKA & 0.1134 & 67.22 & 68.90 \\
\midrule
ResNet-18 $\rightarrow$ ResNet-50 & CKA & 69.66 & 76.54 & 76.12\\
\bottomrule
\end{tabular}
\caption{\textbf{Network of Theseus is robust to reverse replacement}. We can reverse replacement in NoT by converting from an MLP to ResNet-18, converting from an RNN designed for image classification to ResNet-18, or converting from ResNet-18 to ResNet-50. We find that this has little effect on results, indicating that NoT is robust. We emphasize the impact of this finding: we do not find impact on network performance when the original guide network is poorly designed.}
\label{tab:reversal}
\vspace{-2ex}
\end{table}

A potential concern is that representational alignment is harmful when the guide network is poorly chosen. In order to test target network sensitivity to a specific guide network, we add two additional experiments. In the first, we replace linear layers in an MLP with convolutional layers from ResNet-18, effectively reversing our ResNet-18$\rightarrow$MLP experiment. In the second, we repeat this with ResNet-18 with ResNet-50, where we replace a ResNet-18 with several ResNet-50 block. Our goal is to see whether this hurts performance of the resultant ResNet-18 or ResNet-50, specifically by changing geometric aspects of the architecture, leading us to believe that we are more sensitive to guide network architecture.

We report results in \Cref{tab:reversal}. We find that when the guide network is ill-suited for the given task, this does not affect the performance of the target network on the same task, given that the target network is well-suited for the downstream task. This demonstrates that NoT is not overly sensitive to guide network choices and a given CKA target won't derail downstream performance. 
 
\section{Comparison with Progressive Distillation}
\begin{table}[t]
    \centering
    \begin{tabular}{ccccc}
    \toprule
        Setting & Guide & NoT Target & Distillation Target & Original \\ \midrule
        ResNet-18$\rightarrow$MLP & 69.16 & 62.55 & 59.13 & 40.01 \\ 
        Untrained ResNet-18$\rightarrow$MLP & 0.10 & 60.28 & 31.11 & 40.01 \\ \midrule
        GPT-2$\rightarrow$RNN ($\downarrow$) & 37.50 & 50.58 & 75.49 & 119.44 \\ 
        Untrained GPT-2$\rightarrow$RNN ($\downarrow$) & 51948.4 & 58.26 & 135.55 & 119.44 \\ \bottomrule
    \end{tabular}
    \caption{\textbf{Progressive distillation}: We include a comparison to knowledge distillation as a method for aligning our converted architectures to our original architectures. We progressively convert and distill ResNet-18 into our MLP and GPT-2 into our RNN. We only apply this method for conversion where the guide and target have the same layer shapes to allow propagation to the final logits. We find competitive results with NoT when the guide is trained and much worse results when the guide is untrained. This emphasizes that NoT can exploit untrained networks.}
    \label{tab:distillation}
\end{table}

While in this work, we mainly use representational methods to align our converted target architecture to the guide architecture, another method to perform such alignment is knowledge distillation \citep{hinton2015distilling}. We introduce knowledge distillation as part of our progressive replacement schedule as a comparison to NoT, which we refer to as \emph{progressive distillation}. 

At each replacement stage, we convert a guide network layer to a target layer and perform knowledge distillation from the guide network to the newly converted architecture. We train the converted layers using the distillation. Due to needing to pass activations to the final logit layer, we only test settings where the activation shapes are the same across the converted layers in the guide network. 

We show results in \Cref{tab:distillation}. We find that progressive distillation performs competitively with NoT with CKA when the guide architecture is trained but performs worse than the target architecture baseline when the guide architecture is untrained. This is intuitive because distillation doesn't provide access to the internal geometry of the architecture which we exploit with CKA. This further establishes that using representational similarity metrics explains our improvement with untrained architectures. 

One interesting finding we would like to highlight is that we find a significant improvement with NoT in our GPT-2$\rightarrow$RNN setting than over distillation. This implies that NoT overcomes a training problem with RNNs that could not be overcome with distillation. Using the output of a teacher model was not sufficient to prevent stability problems in training an Elman RNN. However, using internal representations of a guide architecture could prevent said problems and achieve stronger results. We believe this has significant implications for RNN training and demonstrates that RNN training stability can be improved by using untrained architectures. 

\section{Automating NoT}
\label{ap:auto}
\begin{table} 
\begin{tabular}{lccc}\toprule
                      & Original & NoT (Trained-EMA) & NoT (Trained-Original) \\ \midrule
ResNet-18 $\rightarrow$MLP & 69.66 & 60.96 & 62.12 \\
DINOv2 $\rightarrow$Patch-MLP & 81.03 & 71.45 & 72.56 \\
GPT-2$\rightarrow$RNN & 37.50    & 48.60             & 50.58\\\bottomrule
\end{tabular}
\caption{\textbf{Automating NoT with EMA and new learning rate schedulers}: We automate NoT by tuning the learning rate and stopping criterion with an EMA-based scheduler.}
\label{tab:auto}
\end{table}

One major component we aim to address is that NoT seems to rely on a hand-designed schedule where every layer has the learning rate and number of epochs of training tuned for optimal representational similarity. To overcome this limitation, we also discuss further methodology for automating the learning rate and number of epochs of training applied per stage of representational similarity optimization. 

Our automation applies an exponential moving average (EMA) over the representational similarity loss at every stage to measure the average change over a 500 step window. We then use this EMA as a learning rate condition and stopping condition. For our learning rate, we start with a consistent setting for all stages. We reduce the learning rate when our EMA plateaus over 4 epochs \citep{pytorch_reducelronplateau}. We then stop our optimization when we saturate the EMA of the representational similarity loss i.e. we reach a CKA loss of less than 0.01 or train for 100 epochs. We find that this threshold is sufficient to get a strong result. 

We apply this to our cross-network architecture conversions including ResNet$\rightarrow$MLP, DINOv2$\rightarrow$Patch-MLP, and GPT-2$\rightarrow$RNN in \Cref{tab:auto}. We convert GPT-2 to an RNN with a starting learning rate of 1e-4 at every stage, allowing the scheduler to tune the learning rate, and using an EMA over 500 steps to measure when to apply the scheduler and when to stop the training. For each stage, we either train until the similarity loss for CKA is less than 0.01 or stop training after 100 epochs. We find that our automatic tuning leads to a slightly stronger result than what was reported for NoT. This reduces the expensive nature of NoT. 

\section{Hidden-State Adapters}
\label{ap:hidden_adapters}

We evaluate two direct alternatives to NoT: hidden-state conversion with adapters and standard logit distillation. For hidden-state conversion, we insert a learned linear adapter that maps target-layer representations to the corresponding guide-layer representations, using either a trained (T) or untrained (U) guide. Due to memory constraints, we did not run the adapter baseline for the ResNet-18$\rightarrow$MLP setting. For logit distillation, we train the target with standard teacher-logit supervision from a trained guide. Results are shown in \Cref{tab:hidden_adapters}. Hidden-state adapters are competitive with NoT in the tested settings, which supports the broader motivation of NoT: representational alignment can transfer computations across architecture families. Unlike a fixed adapter baseline, NoT progressively replaces guide modules with target modules, leaving only the target architecture at inference.

\begin{table}[t]
\centering
\begin{adjustbox}{max width=\textwidth}
\begin{tabular}{lccc}
\toprule
Task & NoT (T/U) & Adapter (T/U) & Logit-T \\
\midrule
ResNet$\rightarrow$MLP (IN-1K) & $\times$ / $\times$ & $\times$ / $\times$ & 46.11 \\
DINOv2$\rightarrow$Patch-MLP (IN-1K) & 71.45 / 72.56 & 69.11 / 67.91 & 69.47 \\
GPT-2$\rightarrow$RNN (WT) & 48.60 / 50.58 & 45.18 / 44.45 & 58.19 \\
\bottomrule
\end{tabular}
\end{adjustbox}
\caption{\textbf{Hidden-state adapters and logit distillation.} We compare NoT with a linear hidden-state adapter and standard logit distillation from a trained guide. T/U denotes trained/untrained guide results. IN-1K reports ImageNet Top-1 accuracy (higher is better), and WT reports Wikitext perplexity (lower is better).}
\label{tab:hidden_adapters}
\end{table}

\section{Accuracy-Efficiency Tradeoff}
\label{ap:efficiency}

We provide inference-time measurements for a representative NoT conversion in \Cref{tab:latency}. The main paper already includes settings with concrete accuracy--efficiency tradeoffs, most notably ResNet-50$\rightarrow$ResNet-18 and GPT-2 Large$\rightarrow$GPT-2 Small, where the target is shallower while improving over the baseline target accuracy. We also measure wall-clock latency for the DINOv2$\rightarrow$Patch-MLP conversion. Using an H100 and an RTX-3090, we measure one forward pass over a single image and report the mean and standard deviation over ten runs. The Patch-MLP target is faster on both devices while preserving much of the guide accuracy reported in the main results.

\begin{table}[t]
\centering
\begin{adjustbox}{max width=\textwidth}
\begin{tabular}{lcccc}
\toprule
Setting & Guide Time (H100) & Target Time (H100) & Guide Time (RTX-3090) & Target Time (RTX-3090) \\
\midrule
DINOv2$\rightarrow$Patch-MLP & 0.69 ms/image (0.05) & 0.61 ms/image (0.03) & 1.03 ms/image (0.055) & 0.87 ms/image (0.031) \\
\bottomrule
\end{tabular}
\end{adjustbox}
\caption{\textbf{Inference latency for DINOv2$\rightarrow$Patch-MLP.} We report mean wall-clock latency for one forward pass, with standard deviation in parentheses, over ten runs on each GPU.}
\label{tab:latency}
\end{table}

These measurements complement the accuracy results by showing that NoT-converted target architectures can yield practical inference-time gains. More broadly, NoT opens a space of architecture-transfer experiments in which targets can be selected for deployment constraints and initialized through representational alignment rather than trained from scratch.

\section{Analyzing DINOv2}
\label{ap:dino}
\begin{figure}[t]
    \centering
    \includegraphics[width=0.8\textwidth]{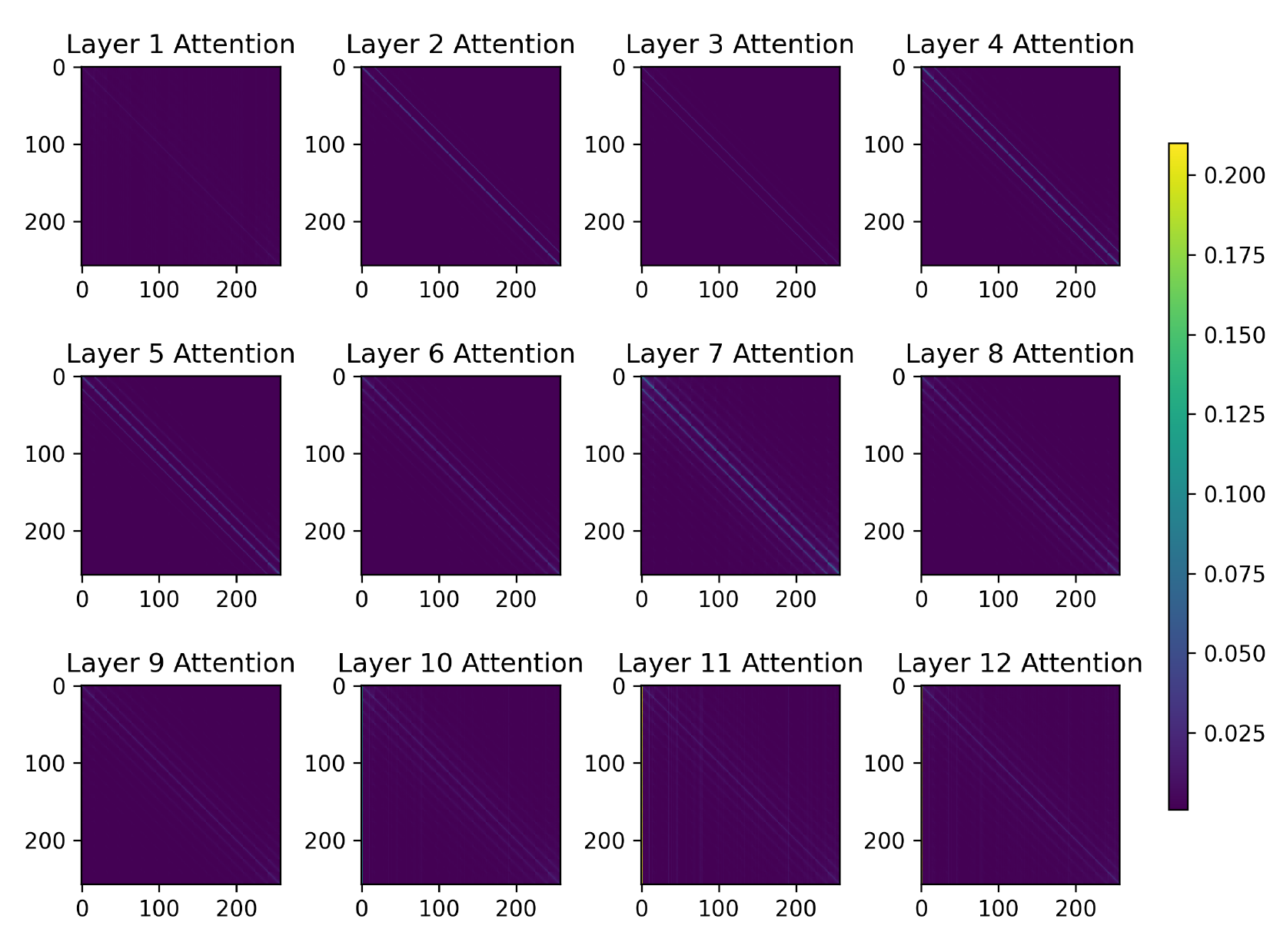}
    \caption{\textbf{DINOv2 has simple attention maps}: We visualize attention maps for DINOv2 over a set of ImageNet images to better understand how we are able to remove cross-token communication via NoT. We find attention is fairly sparse in DINOv2, when focusing on image classification.}
    \label{fig:dinov2_att}
\end{figure}

A surprising result we found was that we could replace attention in DINOv2 with a Patch-MLP. This Patch-MLP has no cross-token communication, with linear layers applying transformations per-token instead. We investigated why we could convert from attention to Patch-MLPs in DINOv2 without much loss in performance. First, we found that the first layer of DINOv2 is a convolution and the last layer aggregates over tokens for a label prediction, meaning that cross-token prediction is not entirely eliminated.

In \Cref{fig:dinov2_att}, we show attention map visualizations for all layers in DINOv2 for a set of ImageNet images. We see that for most DINOv2 layers, attention is highly sparse. Image tokens only focus on tokens in a nearby neighborhood. Most of the other focus goes to the original token itself. Cross-token communication is sparse. We believe this makes attention in DINOv2 much easier to replace with a Patch-MLP. 

This matches intuition in prior findings that attention matrices in ViTs are repetitive across layers \citep{zhang2024you, venkataramanan2023skip} when analyzing attention patterns when completing tasks like image classification. Intuitively, this makes sense. In image classification, there is a central image and cross-token communication may not be entirely necessary. We are not the first to propose removing cross-token computation in ViTs. Other works such as MetaFormer \citep{yu2023metaformer} or RIFormer \citep{wang2023riformer} have proposed methods to replace cross-token mixing in ViTs or MLP-Mixers \citep{tolstikhin2021mlp} due to sparse cross-token communication.

We believe such findings demonstrate the benefits of NoT. We were able to match findings in prior work by directly training a replacement for attention with strong performance. This is indicative that image classification is sparse, requiring a network to focus on a specific token. However, we also believe that such findings may not transfer to other tasks. For example, if we were to focus on semantic segmentation, then we may find that we cannot use NoT for strong performance given the likelihood that semantic segmentation requires cross-token communication. 

\section{D-MNN Analysis}
\begin{figure}
    \centering
    \includegraphics[width=0.9\textwidth]{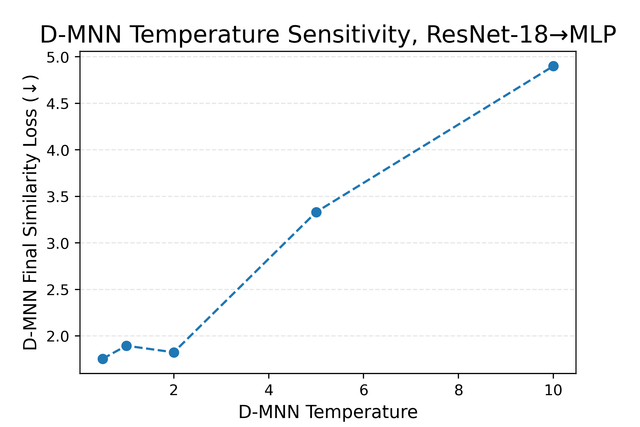}
    \caption{\textbf{Temperature Variations across D-MNN}: We analyze how D-MNN similarity loss varies across different temperature values from 0.5 to 10.0. We see that close temperature values like 0.5 to 2.0 have similarity loss. As we increase similarity we see worse similarity.}
    \label{fig:temp_dmnn}
\end{figure}

We analyze our new similarity metric, D-MNN in \Cref{fig:temp_dmnn}, starting with analyzing how temperature in \Cref{eq:temp} affects our final similarity loss for our ResNet-18$\rightarrow$MLP conversion. We find that close temperature values have similar final similarity losses but as we make larger changes in temperature, we see lower D-MNN similarity between our original and converted network. This is intuitive due to our design of D-MNN, where larger temperatures flatten the probability distribution of the target and guide network representations.
%%%%%%%%%%%%%%%%%%%%%%%%%%%%%%%%%%%%%%%%%%%%%%%%%%%%%%%%%%%%

% \newpage
% \input{checklist.tex}

\end{document}